\pdfoutput=1

\documentclass[11pt]{article}

\usepackage{acl}
\usepackage{times}
\usepackage{latexsym}
\usepackage{hyperref}
\usepackage{xcolor}
\usepackage{soul}
\definecolor{mygold}{HTML}{FFD243}
\definecolor{mygreen}{HTML}{80C535}
\definecolor{myorange}{HTML}{FFC000}
\definecolor{myred}{HTML}{FF576A}
\sethlcolor{mygold}

\usepackage[T1]{fontenc}

\usepackage[utf8]{inputenc}
\usepackage{graphicx}
\usepackage{amsmath,amssymb}
\usepackage[super]{nth}

\usepackage{microtype}
\usepackage{inconsolata}

\title{``Image, Tell me your story!''\\ Predicting the original meta-context of visual misinformation}

\author{
Jonathan Tonglet$^{1,2}$, Marie-Francine Moens$^{3}$, Iryna Gurevych$^{1}$
\\
        \textsuperscript{1}Ubiquitous Knowledge Processing Lab (UKP Lab), \\ Department of Computer Science 
and Hessian Center for AI (hessian.AI),\\ TU Darmstadt \\ 
\textsuperscript{2} Department of Electrical Engineering, KU Leuven\\
\textsuperscript{3} Department of Computer Science, KU Leuven\\ 
\href{https://www.ukp.tu-darmstadt.de}{www.ukp.tu-darmstadt.de}
}

\begin{document}
\maketitle
\begin{abstract}
To assist human fact-checkers, researchers have developed automated approaches for visual misinformation detection. These methods assign veracity scores by identifying inconsistencies between the image and its caption, or by detecting forgeries in the image.
However, they neglect a crucial point of the human fact-checking process: identifying the original meta-context of the image. By explaining what is \textit{actually true} about the image, fact-checkers can better detect misinformation, focus their efforts on check-worthy visual content, engage in counter-messaging before misinformation spreads widely, and make their explanation more convincing. Here, we fill this gap by introducing the task of automated image contextualization. We create 5Pils, a dataset of  1,676 fact-checked images with question-answer pairs about their original meta-context. Annotations are based on the 5 Pillars fact-checking framework. We implement a first baseline that grounds the image in its original meta-context using the content of the image and textual evidence retrieved from the open web. Our experiments show promising results while highlighting several open challenges in retrieval and reasoning. We make our code and data publicly available.\footnote{\href{https://github.com/UKPLab/5pils}{github.com/UKPLab/5pils}} 
\end{abstract}

\section{Introduction}
\label{sec:intro}

\begin{figure}
    \centering
    \includegraphics[scale=0.36]{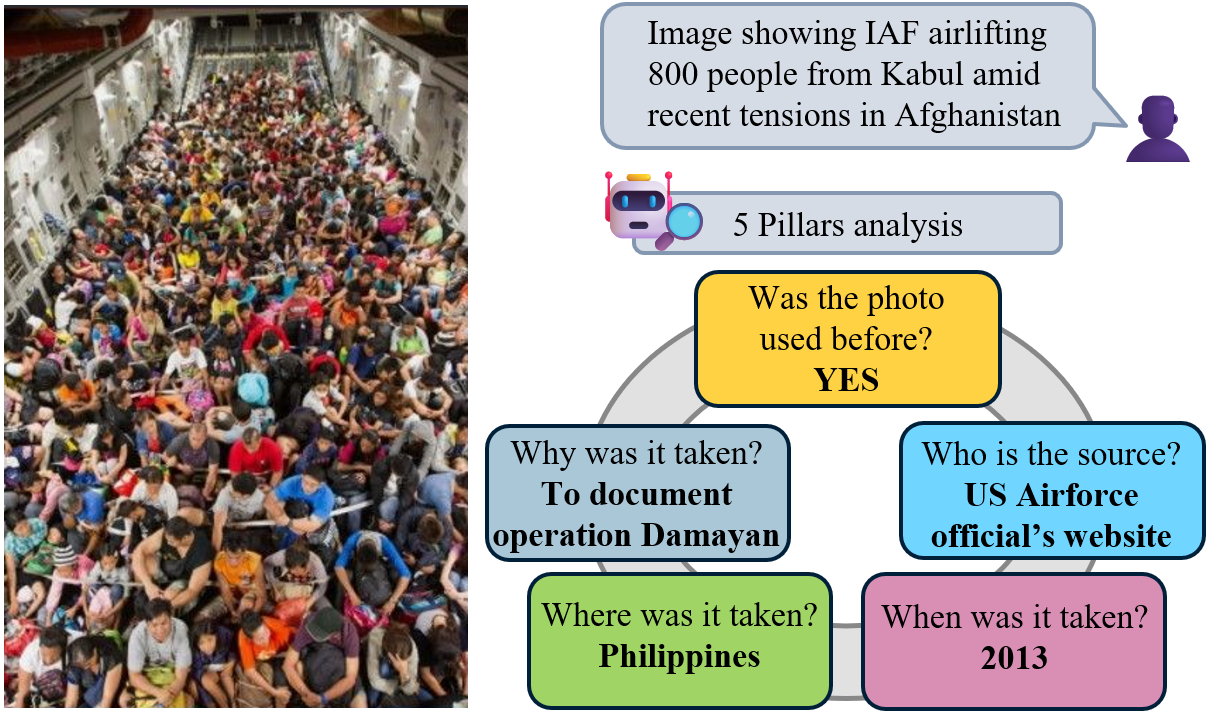}
    \caption{An out-of-context image with a false caption. The original meta-context of the image is established by answering the questions of the 5 Pillars framework.}
    \label{fig:intro}
\end{figure}
The surge of visual misinformation poses a growing threat to our society. In 2023, more than 30\% of all fact-checked claims contained images, an increasing portion of which are AI-generated \citep{dufour2024ammeba}. Journalists and fact-checkers have developed pipelines for detecting and debunking visual misinformation \citep{silverman2013verification,urbani2020verifying,khan2023online,10017287,dufour2024ammeba}. Image contextualization is a key component of these pipelines: it establishes the meta-context of the image by collecting evidence and answering a concise set of questions.
\citet{urbani2020verifying} proposes a framework, which we refer to as ``5 Pillars'', where information about the  Provenance (Was the image used before?), Source (Who is the source?), Date (When was it taken?), Location (Where was it taken?), and Motivation (Why was it taken?) serve as the image meta-context. For example, Figure \ref{fig:intro} shows an out-of-context image purporting to represent an Indian plane evacuating people from Kabul in 2021. After conducting a 5 Pillars contextualization, fact-checkers established that the image was taken in 2013 in a US plane in the Philippines.\\
Image contextualization benefits to several other components of the human fact-checking (FC) pipeline. (1) It helps identify inconsistencies between false claims and images, as shown in Figure \ref{fig:intro}. (2) It is a mandatory component of writing a solid debunking article. Indeed, debunking requires not only stating \textit{why a claim is false} but also \textit{what alternative is true instead} \citep{lewandowsky2020debunking}. (3) It allows writing prebunking articles \citep{guo-etal-2022-survey}, by informing about the image's true meta-context before misinformation spreads widely. (4) It helps assess whether the image is check-worthy. For example, images published by an unreliable source, or depicting critical events (war, crises, ...) deserve more debunking efforts.\\
Image contextualization is a time-consuming task that involves finding several evidence pieces, and combining various tools that require expertise \citep{silverman2013verification}.
Despite its crucial position in the human FC pipeline, the automation of image contextualization has been ignored in favor of other components, predominantly manipulation \citep{huh2018fighting,liu2023unified,Shao_2023_CVPR} and image-caption inconsistency detection \citep{luo-etal-2021-newsclippings,Abdelnabi_2022_CVPR,papadopoulos2024verite,Qi_2024_CVPR}. The output of these tasks is a score, indicating whether or not the content is misinformation. Fact-checkers have expressed skepticism about the real-world performance of these scores, with the risk of publishing incorrect automated verdicts \citep{arnold2020fullfact}. In contrast, there is a real interest in methods that provide evidence and additional context \citep{arnold2020fullfact,ijcai2021p619,schlichtkrull-etal-2023-intended}, which includes image contextualization.\\
The contributions of this work are as follows: (1) We fill the gap by introducing the novel task of automated image contextualization. Given an image, a model must identify its meta-context using the image content and evidence from the open web.
(2) We collect 5Pils, the first dataset with question-answer pairs about the image meta-context based on the 5 Pillars framework \citep{urbani2020verifying}. (3) We create the first baseline for image contextualization with textual evidence retrieved from the open web. Our extensive quantitative and qualitative analyses reveal that the task is challenging, even for SOTA large language models (LLMs). Following \citet{schlichtkrull-etal-2023-intended},  Appendix \ref{sec:epistemic} presents the intended uses of this work.

\section{Related work}

\textbf{Visual misinformation}\\ 
 Existing work on visual misinformation focuses on assigning a veracity score to images. Many real-world \citep{zlatkova-etal-2019-fact,suryavardan2023factify,papadopoulos2024verite} and synthetic \citep{jaiswal2017multimedia,10.1145/3372278.3390670,luo-etal-2021-newsclippings,biamby-etal-2022-twitter,10.1145/3592572.3592842} datasets have been introduced. Separate methods are used for \textit{out-of-context} images, where a score is assigned based on image-caption inconsistencies \citep{khattar2019mvae,9378019,9746193}, and \textit{manipulated} and \textit{fake} images, where forgeries are detected in the image \citep{huh2018fighting,liu2023unified,Shao_2023_CVPR}. Some works rely on external evidence \citep{zlatkova-etal-2019-fact,Abdelnabi_2022_CVPR,pham2023detecting,10.1145/3539618.3591896,papadopoulos2023red,yuan-etal-2023-support,10.1145/3581783.3612183,Qi_2024_CVPR,papadopoulos2024similarity}, while others use only the image and the caption \citep{nakamura-etal-2020-fakeddit,luo-etal-2021-newsclippings,aneja2023cosmos,papadopoulos2024verite,9746193}.
 \citet{akhtar-etal-2023-multimodal} provides a comprehensive survey of the field.
 Unlike thes  works, we use the image and evidence not to provide a veracity score, but to ground the image in its original meta-context.\\
 
\noindent \textbf{Question Answering for Fact-Checking} \\
Representing automated FC (AFC) as a question answering (QA) task has attracted recent interest, especially for text claims that require several reasoning steps \citep{chen-etal-2022-generating,ousidhoum-etal-2022-varifocal,yang2022explainable,pan-etal-2023-fact,NEURIPS2023_cd86a305}. By decomposing the reasoning, the verdict becomes more human-interpretable \citep{yang2022explainable,rani-etal-2023-factify}. \citet{rani-etal-2023-factify} and \citet{chakraborty-etal-2023-factify3m} introduce a framework for decomposing a text claim in atomic statements, which they call the 5W (who, what, when, where, and why). Unlike our work, these approaches do not ask questions about images, nor do they try to identify the meta-context of the image.

\section{The 5 Pillars of Image Verification}

The 5 Pillars of image verification is a framework proposed by \citet{urbani2020verifying}, which summarizes years of insights on journalistic best practices in image verification \cite{silverman2013verification,urbani2020verifying,bellingcat2021,khan2023online}. It divides image contextualization into 5 components, as shown in Figure \ref{fig:intro}: the Provenance, the Source, the Date, the Location, and the Motivation. \\
\textbf{Provenance} answers the question ``\textit{Was the image used before?}'' \citep{khan2023online}. Identifying prior uses of the image is essential for answering the subsequent questions. Hence, it is considered the most important pillar \citep{urbani2020verifying}. To answer this question, fact-checkers rely mainly on reverse image search (RIS) engines. These engines take images as input and return (partially) matching images on the Internet \citep{urbani2020verifying}.\\
The \textbf{Source} is the person who captured or created the image, to be distinguished from the person who shared the visual misinformation online. Establishing the Source allows to evaluate the credibility of the original image. A combination of social media profiling, interviews, and RIS allows fact-checkers to answer this question. \\
The \textbf{Date}, i.e., ``\textit{When was the image taken?}'', and the \textbf{Location}, i.e., ``\textit{Where was the image taken?}'', position the image in time and space, and highlight any mismatch with the claimed context. The tasks associated with these two pillars are also known as chronolocation and geolocation \citep{bellingcat2021,khan2023online}. In the simplest case, the image contains the Date in EXIF format. However, most social media platforms remove EXIF data from uploaded images \citep{urbani2020verifying}.  RIS results can also help to find the Date and the Location. If that is not enough, fact-checkers can define keywords to query search engines and retrieve (partially) matching images, a process detailed in Appendix \ref{sec:examples_human_fc}. In other cases, they search for cues in the image, for example, landmarks, road signs, and clothing. Additional tools include street imagery,  sun angle analysis, and historical weather records \citep{urbani2020verifying,bellingcat2021}.\\
The \textbf{Motivation}, asks the question ``\textit{Why was the image taken?}''. In most cases, the Motivation is to report on an event. However, some sources may have a particular agenda, for example, if they are political activists. In other cases, a fake image used for malicious purposes might originate from the harmless work of an artist.  The source's Motivation is different from the claimant's intent, which is explored in \citet{da-etal-2021-edited}. \\
Figure \ref{fig:intro} shows the 5 Pillars analysis of an out-of-context image. Examples with manipulated and fake images are shown in Appendix \ref{sec:examples}. Given its popularity and comprehensive documentation \citep{urbani2020verifying,bellingcat2021,khan2023online}, we use the 5 Pillars framework to define QA pairs for automated image contextualization.

\section{Task definition and evaluation}
\begin{figure*}
    \centering
    \includegraphics[scale=0.34]{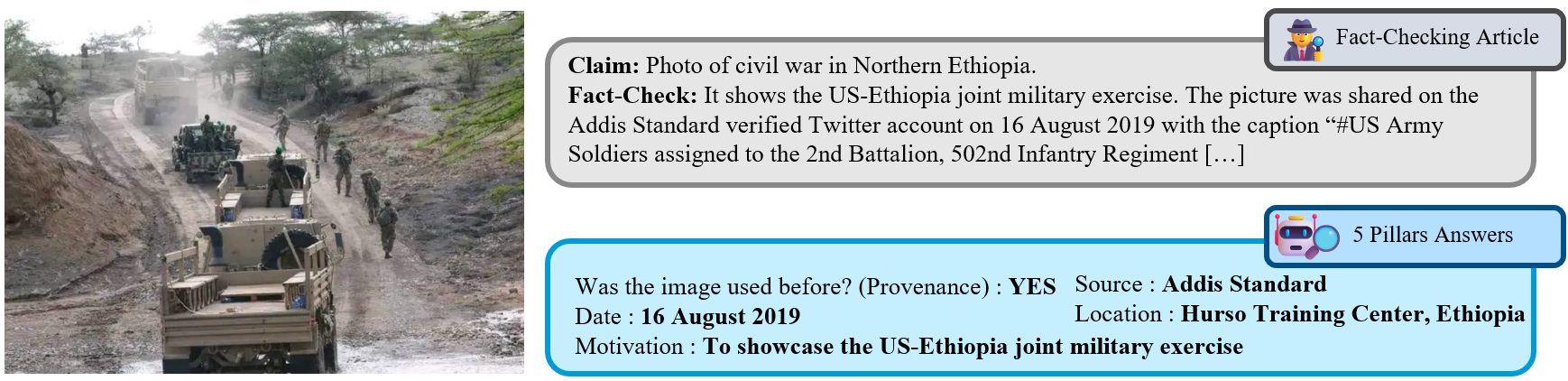}
    \caption{An annotated image of the 5Pils dataset. Above: the FC article. Below: the extracted 5 Pillars answers.}
    \label{fig:5P_example}
\end{figure*}

The task is to provide answers for each of the 5 Pillars using the image content itself and evidence retrieved from the open web. Formally, a model has to generate answer $A$ given a question $Q$, an image $I$, and evidence items $E$ retrieved from the open web $W$ using a retriever $R$,
\begin{equation*}
\resizebox{\columnwidth}{!}{
    $A = \underset{a}{argmax} \: \: P\left(a \mid Q,I,E\right)\text{,} \:\:\: E=R(W)\text{.}$}
\end{equation*}
We impose two restrictions on the open web retriever.
(1) It cannot return evidence published after the FC article from which we collected the image and annotations, to avoid temporal leaks \citep{glockner-etal-2022-missing}. (2) It cannot return FC articles published by other FC organizations.\\
The answer to \textbf{Provenance} (``Was the image used before?'') is either ``Yes'' when (partially) matching images are retrieved, and ``No'' or ``Unknown'' otherwise, the latter two being difficult to distinguish in practice. The answer is automatically derived from the retrieved results. Therefore, the Provenance evaluates the ability of retrievers to obtain evidence that contains previous versions of the same image.  We define the Provenance score as the recall on images with ground truth answer ``Yes''. We do not penalize nor reward retrievers for collecting matching images when the ground truth is ``No'' or ``Unknown''.\\
\textbf{Source} expects free-form answers, evaluated using the RougeL \citep{lin-2004-rouge} and Meteor \citep{banerjee-lavie-2005-meteor} metrics.\\
The expected answers for \textbf{Date} are actual timestamps. We evaluate the answers using the Exact Match (EM) and $\Delta$,  a custom metric ranging from 1 for predictions that are close in time to the ground truth to 0 for those that are far. If the ground truth answer contains more than one date for the image, for example, if it is a composite image, we compute  $\Delta$ based on the smallest time difference that can be obtained by matching pairs of predicted and ground truth answers. Formally $\Delta$ is defined as 
 \begin{equation*}
    \Delta = \frac{1}{N} \sum_{(i,j) \in M} \frac{1}{1 + d_{i, j}}\:\text{,}
\end{equation*}
where $N$ is the number of ground truth elements, $M$ is the set of matched  prediction-ground truth pairs $(i,j)$, and $ d_{i, j}$ is the optimal distance between $i$ and $j$, obtained with the Jonker-Volgenant algorithm \citep{jonker1987shortest,7738348}. Years serve as the distance unit. \\
\textbf{Location} is evaluated with RougeL and Meteor. The subset of images where ground truth locations can be mapped to coordinates using GeoNames\footnote{\href{https://www.geonames.org/}{geonames.org}} is further evaluated in a spatial dimension. We employ variants of $\Delta$, namely, Coordinates $\Delta$ (CO$\Delta$), with thousands kilometers as distance unit, and Hierarchical Location $\Delta$ (HL$\Delta$), using the distance between the prediction and ground truth in terms of their GeoNames hierarchies.\footnote{\href{https://www.geonames.org/export/place-hierarchy.html}{geonames.org/export/place-hierarchy.html}} For example, the distance for the pair (USA, Chicago) is 2, based on the hierarchy \textit{USA-Illinois-Chicago}.\\
In addition to RougeL and Meteor, we evaluate \textbf{Motivation} with BertScore (BertS) \citep{bert-score} to measure the semantic similarity between ground truth and predicted Motivation.

\section{The 5Pils Dataset}

The 5Pils dataset consists of 1,676 images annotated with 5 Pillars answers. We collect metadata for analysis purposes, including the \textit{type of image}, the \textit{verification strategy}, and the \textit{verification tools}.

 \subsection{Data Collection}

 We selected FC organizations that met the following criteria: they write some of their articles in English, include the verified image in the article, do not systematically cover the images with FC annotations, for example, marking the image with a ``Fake'' stamp or a red cross, and are verified signatories of the International Fact-Checking Network (IFCN)\footnote{\href{https://ifcncodeofprinciples.poynter.org/signatories}{ifcncodeofprinciples.poynter.org/signatories}} since Fall 2023, which guarantees that the images are verified according to the highest standards. None of the existing visual misinformation datasets met these combined criteria. However, three IFCN members did: Factly, PesaCheck, and 211Check, based in India, Kenya, and South Sudan, respectively. Hence, we created a new dataset with articles from these organizations. We collected all URLs archived on the Wayback Machine between 2019 and 2023.\footnote{\href{https://archive.org/help/wayback_api.php}{archive.org/help/wayback\_api.php}}  We kept URLs that contain the keywords \textit{photo}, \textit{image}, or \textit{picture}. This results in a collection of 3,424 articles. We scraped the \textit{title}, the \textit{publication date}, the \textit{text content}, and the fact-checked \textit{image URL}. 
 We downloaded the images and cleaned them by cropping social media sidebars. We applied a second round of filtering by removing articles written in French, Kannada, and Telugu, articles that do not verify images despite the keyword match, articles duplicated from one organization to another, and articles with FC annotations covering the image. The resulting unlabeled dataset contains 1,692 FC articles.

\begin{figure*}[!ht]
\centering
\includegraphics[scale=0.42]{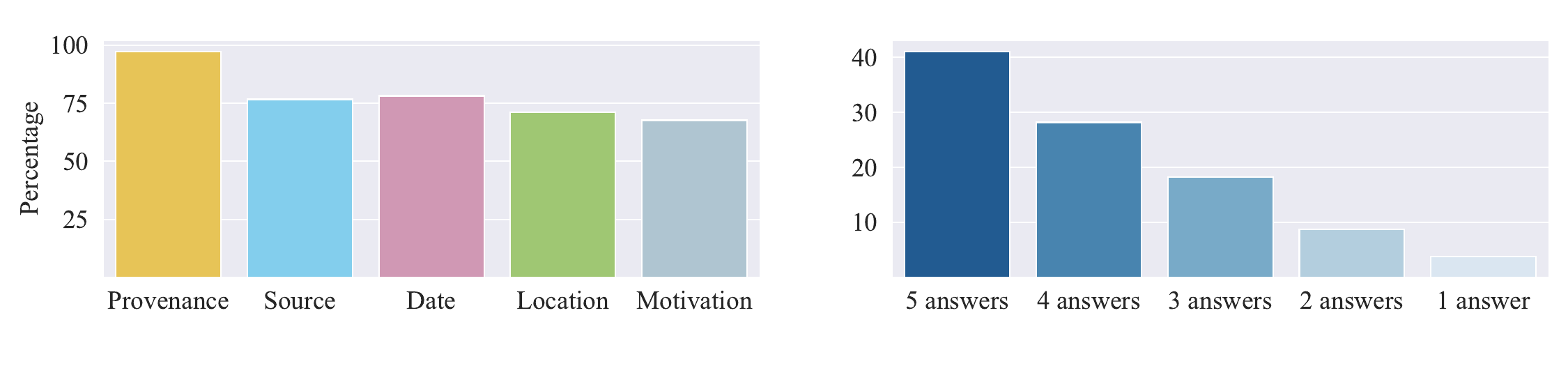}
\caption{Left: Images with answers per pillar (\%). Right: Images with n answered pillars (n=1,..,5) (\%).}
\label{fig:label_stats}
\end{figure*}

\subsection{Data Annotation}

We make the assumption that each article explicitly discusses one or more of the 5 Pillars. Hence, the annotation process becomes a text extraction task. Given its relative simplicity, we decided to automate the process using GPT4 \citep{openai2023gpt4}. We tasked GPT4 to extract the 5 Pillars answers and relevant metadata as a JSON file using a custom prompt detailed in Appendix \ref{sec:prompt}.  An annotated image is illustrated in Figure \ref{fig:5P_example}. In total, only 16  articles (\textless1\%) did not contain any 5 Pillars answers, confirming our initial assumption. We recruited three students to evaluate the quality of the GPT4 annotations for a random sample of 50 images. After reading the corresponding FC article, they assigned a label to each of the GPT4 extracted answers:  ``Correct'' if GPT4 detects the presence or absence of the answer in the article, ``Missing'' if GPT4 does not provide an answer while it was available in the article, and ``Incorrect'' if GPT4 provides a wrong answer.
Taking their majority vote, annotators found 100, 100, 98, 96, and 94\% of the answers to be ``Correct'' for Provenance, Source, Date, Location, and Motivation, respectively, the remaining cases being ``Missing'' answers. Restricting to a unanimous vote for ``Correct'' answers, those percentages decrease to 94, 82, 90, 82, 76\%. For the rest, at least one annotator has chosen the ``Missing'' (2, 10, 10, 12, 20\%) or ``Incorrect'' (4, 8, 0, 6, 4\%) label. ``Missing'' are preferable over ``Incorrect'' labels as they lower the recall without impacting the precision of the annotations. Overall, the high values of majority and unanimous vote for the ``Correct'' label confirm the quality of the automated annotations. To evaluate inter-annotator agreement, we use Randolph's $\kappa$ \citep{randolph2005free}, a variant of Fleiss' $\kappa$ \citep{fleiss1971measuring} that is better suited for annotations that are heavily skewed towards one label \citep{warrens2010inequalities,NEURIPS2023_cd86a305}, ``Correct'' in this case. The Randolph's $\kappa$ for each pillar is 0.94, 0.82, 0.87, 0.82, 0.76.  Those scores show a strong inter-annotator agreement.

\subsection{Dataset Statistics}

5Pils contains 1,676 annotated images. This makes it slightly larger than other real-world visual misinformation datasets \citep{zlatkova-etal-2019-fact,papadopoulos2024verite}.
 We split the dataset into a train, validation, and test set (60\%, 10\%, 30\%). The split is performed along the time dimension to limit temporal leakage \citep{glockner-etal-2022-missing,NEURIPS2023_cd86a305}. The sequential end dates are May 31$^{st}$ 2022, Sep. 20$^{th}$ 2022, and Dec. 28$^{th}$ 2023. Figure \ref{fig:label_stats} shows the 5Pils answers statistics. All pillars are present in 68 \% or more of the FC articles. Furthermore, answers are often present in combination, with 87\% of the images having 3 or more answered pillars. These high values confirm the prevalence of image contextualization and the 5 Pillars approach in human FC practices. Appendix \ref{sec:cooccurence} reports the co-occurrence of 5 Pillars answers. We searched for keywords in the FC articles referring to common strategies and tools for image contextualization. 86\% of the FC articles explicitly refer to RIS, 1.94\% to geolocation techniques, and 1.9\% to keyword search. Google Image Search is the most popular RIS engine (36.7\%), followed by Yandex Images (5.2\%), and Bing Visual Search (3.8\%). Additional analysis about the type of misinformation, the source, temporal, and spatial distributions of images is provided in Appendix \ref{sec:misinfo_type}, \ref{sec:source_analysis}, \ref{sec:temporal}, and \ref{sec:geographic}, respectively.

\section{Experiments}

\subsection{Baseline model}

\begin{figure*}
    \centering
    \includegraphics[scale=0.45]{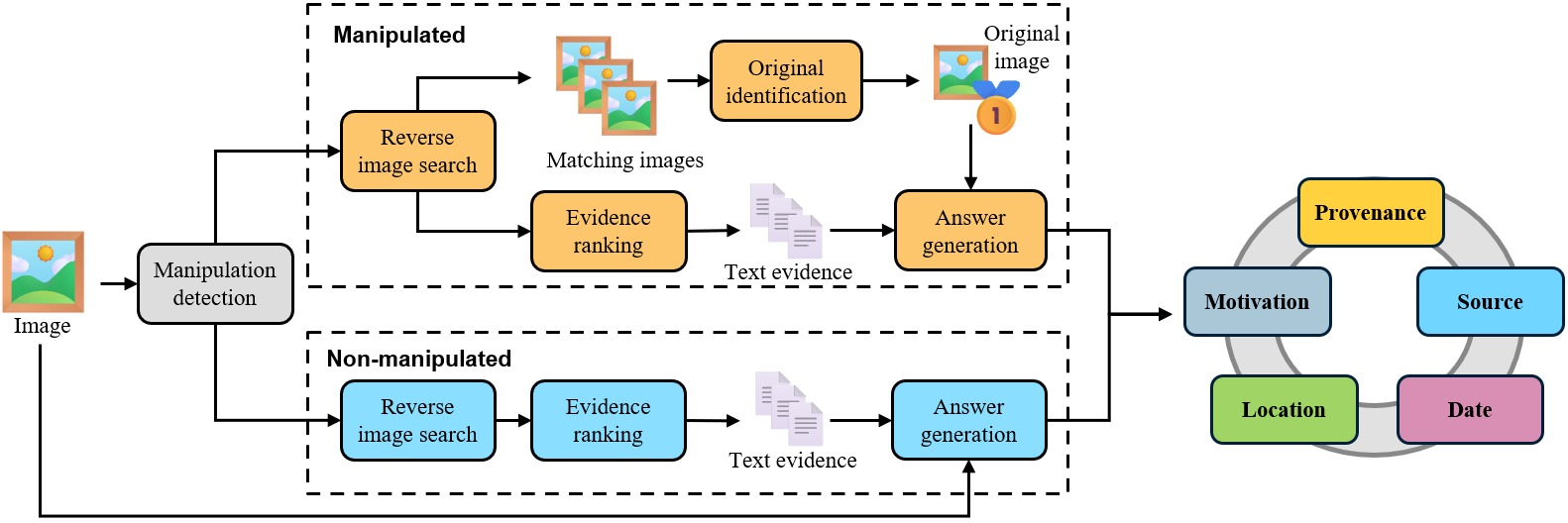}
    \caption{Our baseline for image contextualization. The retriever is a RIS engine that searches web-pages containing previous versions of the image. The top $k$ evidence and the image are provided as input to a QA model to answer the 5 Pillars questions.  Manipulated images go through the additional step of identifying the original unaltered image.}
    \label{fig:baseline}
\end{figure*}

To our knowledge, there is no prior method for automated image contextualization. Therefore, we introduce a first baseline for the task,  illustrated in Figure \ref{fig:baseline}. The baseline is a pipeline based on the best practices of human fact-checkers and the existing AFC literature \citep{akhtar-etal-2023-multimodal}.\\
\textbf{Manipulation detection} To detect images that have been manipulated, we fine-tune a vision transformer (ViT) \citep{DBLP:conf/iclr/DosovitskiyB0WZ21} on the train set. The input is the image and the target is the \textit{type of image} metadata (manipulated, non-manipulated).\\
\textbf{Evidence retrieval} Collecting evidence from the open web requires a retriever. Given its popularity among human fact-checkers, we rely on the Google RIS engine.\footnote{\href{https://cloud.google.com/vision/docs/detecting-web}{cloud.google.com/vision/docs/detecting-web}} We provide the image as input and obtain up to 37  web-pages, 6.84 on average, that used (partially) matching images. Each web-page acts as a single text evidence. We scrape their content with Trafilatura \citep{barbaresi-2021-trafilatura}.\footnote{\href{https://trafilatura.readthedocs.io/en/latest/}{trafilatura.readthedocs.io}}\\
\textbf{Evidence Ranking} We select the top $k$=3 text evidence, which corresponds, on average, to selecting half the evidence matching a given image. We sort the evidence by the cosine similarity between the embedding of the image to verify and the text evidence embedding computed with CLIP \citep{pmlr-v139-radford21a}.\\
\textbf{Original identification} This step is applied to manipulated images to identify the original, unaltered image among the RIS results. The intuition is that the alterations might have removed key elements to identify the original meta-context. As a simple heuristic, we select the retrieved image with the oldest publication date.\\ 
\textbf{Answer generation} We generate zero-shot answers with the following LLMs: Llama2 7B-Chat \citep{touvron2023llama}, Llava 1.5 7B \citep{NEURIPS2023_6dcf277e,Liu_2024_CVPR}, and GPT4 \citep{openai2023gpt4}. We try different input modalities: only the image, only the text evidence, or both. Llama2 is only used with text evidence. For the multimodal setting, we experiment with few-shot predictions using 1 or 2 \textbf{demonstrations}.   The prompt template and implementation details are provided in Appendix \ref{sec:prompts} and \ref{sec:details},  respectively.

\subsection{Main Results}

\begin{table*}
\centering
\resizebox{\textwidth}{!}{ %
\begin{tabular}{cccccccccccccccc}
\hline
 &  & \multicolumn{2}{c}{\textbf{Source}} & & \multicolumn{2}{c}{\textbf{Date}} & & \multicolumn{4}{c}{\textbf{Location}} &  & \multicolumn{3}{c}{\textbf{Motivation}} \\
\textbf{Method} & \textbf{N} &  \textbf{RougeL} & \textbf{Meteor} &  & \textbf{EM} & \textbf{$\Delta$} & & \textbf{RougeL} & \textbf{Meteor} &  \textbf{HL$\Delta$} &   \textbf{CO$\Delta$}  & & \textbf{RougeL} & \textbf{Meteor} & \textbf{BertS}  \\ 
\hline
 & \multicolumn{14}{c}{\textbf{Image only}} & \\
 \hline
Llava  & 0 & 1.16$_{\pm0.2}$ & 0.67$_{\pm0.1}$ &  &  0.00$_{\pm0.0}$ & 13.49$_{\pm0.3}$  & & 5.79$_{\pm0.2}$ & 3.79$_{\pm0.2}$   &  12.74$_{\pm0.5}$ &  16.99$_{\pm0.8}$ & & 7.12$_{\pm0.3}$ & 3.33$_{\pm0.2}$ & 68.5$_{\pm0.1}$ \\
GPT4 & 0 & 2.78$_{\pm0.0}$ & 2.37$_{\pm0.1}$ & & 0.17$_{\pm0.1}$ & 5.68$_{\pm2.2}$ & & 15.73$_{\pm0.3}$ & 15.68$_{\pm0.2}$  & 22.15$_{\pm0.5}$ & 27.31$_{\pm0.6}$ &  & 14.81$_{\pm0.1}$ &  11.08$_{\pm0.7}$  & 65.9$_{\pm0.4}$  \\
\hline
 & \multicolumn{14}{c}{\textbf{Text evidence only}} & \\
 \hline
Llama2  & 0 & 2.83$_{\pm0.0}$  & 4.47$_{\pm0.0}$ &  & 6.06$_{\pm0.0}$ & 37.73$_{\pm0.0}$ & & \textbf{28.65}$_{\pm0.0}$ & \textbf{29.65}$_{\pm0.0}$  & 39.96$_{\pm0.0}$ & 44.81$_{\pm0.0}$ &  & 14.89$_{\pm0.0}$ & 15.65$_{\pm0.0}$ & 52.6$_{\pm0.0}$\\  
GPT4 & 0 & 4.18$_{\pm0.0}$ & 3.03$_{\pm0.0}$ &  & \textbf{7.58}$_{\pm0.0}$ & 23.71$_{\pm0.2}$ & & 24.95$_{\pm3.1}$ &  23.63$_{\pm3.0}$  & 28.04$_{\pm3.1}$  & 31.35$_{\pm2.9}$ & & 15.23$_{\pm0.2}$ & 12.58$_{\pm0.0}$ & 51.0$_{\pm0.4}$\\ 
\hline
 & \multicolumn{14}{c}{\textbf{Image + Text evidence}}&  \\
 \hline
 & 0 & 6.73$_{\pm0.1}$ & 5.00$_{\pm0.1}$ &  &  5.06$_{\pm0.4}$ & \textbf{39.42}$_{\pm0.1}$ & & 25.52$_{\pm0.3}$ & 19.48$_{\pm0.3}$   & 33.08$_{\pm0.4}$ & 40.32$_{\pm0.5}$ &  & 8.71$_{\pm0.1}$ &  4.21$_{\pm0.1}$ & 69.0$_{\pm0.1}$\\ 
  Llava      & 1 & 6.69$_{\pm0.2}$  & 5.08$_{\pm0.1}$  &  & 4.22$_{\pm0.3}$  & 34.96$_{\pm0.2}$  & & 25.63$_{\pm0.6}$  & 21.53$_{\pm0.6}$   & 37.06$_{\pm0.7}$  & 44.92$_{\pm0.5}$  &  & 19.25$_{\pm0.1}$  & 14.09$_{\pm0.2}$  & \textbf{74.6}$_{\pm0.1}$\\
        & 2 & 5.96$_{\pm0.2}$ & 4.65$_{\pm0.1}$ &  & 4.86$_{\pm0.4}$ & 37.67$_{\pm0.1}$ & & 26.14$_{\pm0.5}$& 21.79$_{\pm0.3}$  &  \textbf{40.27}$_{\pm0.4}$ & \textbf{49.52}$_{\pm0.6}$&  &19.33$_{\pm0.3}$ & 13.36$_{\pm0.2}$ & 74.7$_{\pm0.1}$\\

 & 0 & 6.54$_{\pm0.3}$ & 5.31$_{\pm0.4}$ &  & 6.31$_{\pm0.2}$ & 19.92$_{\pm0.7}$ & & 27.63$_{\pm0.6}$ & 26.71$_{\pm0.4}$   & 32.20$_{\pm0.8}$  & 36.82$_{\pm0.9}$ & & 18.58$_{\pm0.3}$ & 15.53$_{\pm0.1}$ & 67.1$_{\pm0.0}$\\ 
GPT4 & 1  & 7.19$_{\pm0.2}$ & 5.61$_{\pm0.2}$ &  & 6.23$_{\pm0.1}$ & 21.32$_{\pm1.4}$ & & 28.15$_{\pm0.5}$ & 26.47$_{\pm0.6}$  & 31.59$_{\pm0.3}$ & 36.22$_{\pm0.4}$ & & 19.17$_{\pm0.1}$ & 16.58$_{\pm0.0}$ & 67.4$_{\pm0.2}$\\ 
  & 2 & \textbf{8.06}$_{\pm0.1}$ & \textbf{6.46}$_{\pm0.2}$ &  & 5.56$_{\pm0.3}$ & 19.00$_{\pm0.1}$ & & 27.34$_{\pm1.5}$ & 26.34$_{\pm1.6}$   & 30.57$_{\pm1.3}$  & 34.81 $_{\pm1.5}$ & & \textbf{19.96}$_{\pm0.2}$ & \textbf{17.02}$_{\pm0.0}$ & 66.9$_{\pm0.4}$\\ 
\hline

\end{tabular}}
\caption{Zero- and few-shot results for answer generation (\%). \textbf{N} is the number of demonstrations. Reported results are averages over 3 iterations with standard deviations. The best scores are marked in \textbf{bold}.}
\label{tab:results}
\end{table*}

We evaluate our baseline on the 5Pils test set. We do not recruit human experts to compare their performance with our baseline directly. Still, we assume their performance to be near 100\%, as all the image annotations were derived from the investigations of such human experts.\\
The \textbf{Provenance} score is 66.2\%, meaning that the Google RIS engine retrieved scrapable web-pages with (partially) matching images for less than 2 out of 3 images for which matching images are known to be available. We attribute this low score to the use of a single RIS engine. In the absence of retrieved results, human fact-checkers typically use additional RIS engines or rely on keyword search, which we do not incorporate in this baseline.\\
Table \ref{tab:results} shows the results obtained for the \textbf{Source}, \textbf{Date}, \textbf{Location}, and \textbf{Motivation}. The best models achieve a promising performance, especially for Location and Motivation, with RougeL and Meteor close to or above 20\%. Source performance is much lower. Predicting the Source can rarely be done with the image alone. Hence, it depends on the quality of the evidence ranking and retrieval. The latter is limited by the use of a single RIS engine. Obtaining a high EM for Date is challenging, especially when a specific day or month is expected. The best EM improves from 7.58 to 13.89 and 19.78 if we evaluate the match at the month and year level, respectively. There is more potential to improve $\Delta$. The best value is 39.42\%, or an average distance of one year and a half between the ground truth and the prediction. The best CO$\Delta$ and HL$\Delta$ scores for Location are promising too, indicating an average distance of a thousand kilometers and an average hierarchical distance of 1.5, respectively. 
These results confirm that image contextualization can be automated to some extent. At the same time, they inform us that a simple baseline that only considers RIS engines as open web retrievers and one-step answer generation with LLMs is not sufficient to solve the task.

\begin{table}
    \centering
    \resizebox{\columnwidth}{!}{ %
    \begin{tabular}{ccccc}
       \hline
         & \textbf{Source} & \textbf{Date} & \textbf{Location} & \textbf{Motivation} \\
         \hline
    Time   & 40.15 & 61.9 & 55.75  & 64.18 \\
    CLIP & \textbf{43.55}& \textbf{62.17}   &  \textbf{61.4} & \textbf{71.92} \\
    \hline
    \end{tabular}}
    \caption{nDCG of different ranking methods (\%). Time is the chronological ranking by publication date.}
    \label{tab:ranking_results}
\end{table}

\begin{table}
\centering
\resizebox{\columnwidth}{!}{ %
\begin{tabular}{ccccccccccccccc}
\hline
 &   \multicolumn{2}{c}{\textbf{Source}} & & \multicolumn{2}{c}{\textbf{Date}} & & \multicolumn{4}{c}{\textbf{Location}} &   & \multicolumn{3}{c}{\textbf{Motivation}} \\
\textbf{Method}  &  \textbf{R} & \textbf{M} & & \textbf{EM} & \textbf{$\Delta$} & & \textbf{R} & \textbf{M} &  \textbf{HL$\Delta$} &   \textbf{CO$\Delta$}  & & \textbf{R} & \textbf{M} & \textbf{B} \\ 
 \hline
No detector & 6.3  & 5 & & 8.3 & 21.8 & & 42.6 & 39.7 &  \textbf{43.1} &  \textbf{49.1} & & 20  & 18.5 & 74.7 \\
Detector & 6 & 4.7 & & 8.3 &  23 & & 44.7 & 39.8 &  \textbf{43.1} &  \textbf{49.1} & & 20.8 & 19& 74.7 \\
Perfect detector  & 6.3 & 4.7 & & \textbf{10} & \textbf{24} & & 42.9 & 39.6 &  \textbf{43.1} &  \textbf{49.1} & & 21.6  &  19.7 &  74.7 \\
Oracle & \textbf{9.3} & \textbf{6.5} & & \textbf{10} & 22.7 & & \textbf{45.8} & \textbf{40.4} &  40.6 & 47.6 & & \textbf{22.2}  &  \textbf{20.4} & \textbf{74.8}\\
\hline

\end{tabular}}
\caption{Zero-shot multimodal GPT4 results for the manipulated images. R is RougeL, M is Meteor, and B is BertS.}
\label{tab:results_manipulated}
\end{table}

\subsection{Quantitative analysis}

\textbf{How do different LLMs compare on the task?} For image input, GPT4 slightly outperforms Llava. The difference between Llama2 and GPT4 is smaller for text input. Llama2 even outperforms GPT4 for Location. GPT4 and Llava achieve similar performance with multimodal input. Llava and Llama2 score better $\Delta$s than GPT4 because they always provide an estimate while GPT4 often refrains from answering.\\
\textbf{What is the preferred input modality?} For Source and Motivation, a multimodal input gives better results. For Date and Location, a multimodal input is equal or worse to text evidence, implying that the multimodal LLMs have limited abilities to reason about the dates and locations of images, and their visual understanding can even override correct information present in the text evidence.\\ 
\textbf{What is the impact of demonstrations?} Demonstrations improve the performance for the Source with GPT4, but decrease it with Llava. They help Llava for Location and Motivation by providing additional guidance on the expected type of answer. Their impact for Date is negative for both models. A manual investigation reveals that the models replace some correct zero-shot Date answers with one of the demonstration's answers.\\
\textbf{How performant is the evidence ranking?}
We use as a target the ranking that sorts evidence according to their n-gram overlap with the ground truth answer. We evaluate the ranking with the normalized discounted cumulative gain (nDCG) metric \citep{jarvelin2002cumulated}.
We compare the CLIP ranking with a baseline that sorts evidence by chronological order of publication (Time ranking). The CLIP ranking outperforms the Time ranking for all pillars, as shown in Table \ref{tab:ranking_results}.\\
\textbf{What is the impact of the separated pipeline for manipulated images?}
Applying the fine-tuned ViT and the publication date heuristic, we retrieve a candidate unaltered version for 58 images. With multimodal zero-shot GPT4, little to no gains are observed over a pipeline that ignores these  steps (\textit{No detector}), as shown in Table \ref{tab:results_manipulated}. The results are slightly better with a perfect accuracy manipulation detector (\textit{Perfect detector}), except for Source and Location. The RIS results do not often contain the unaltered image, partly explaining these minor performance gains. As an upper bound, we consider an \textit{oracle} which provides  the original versions of the image reported in the FC article, if available online, and observe a larger improvement, especially for Source and Location. This indicates that although the current pipeline is imperfect, it is meaningful to develop methods that replace a manipulated image with the original, unaltered version. Future work should consider alternatives to RIS for retrieving the original version, or image editing models to remove manipulated content, obtaining an image close to the unaltered version. 

\begin{figure*}[!ht]
    \centering
    \includegraphics[scale=0.55]{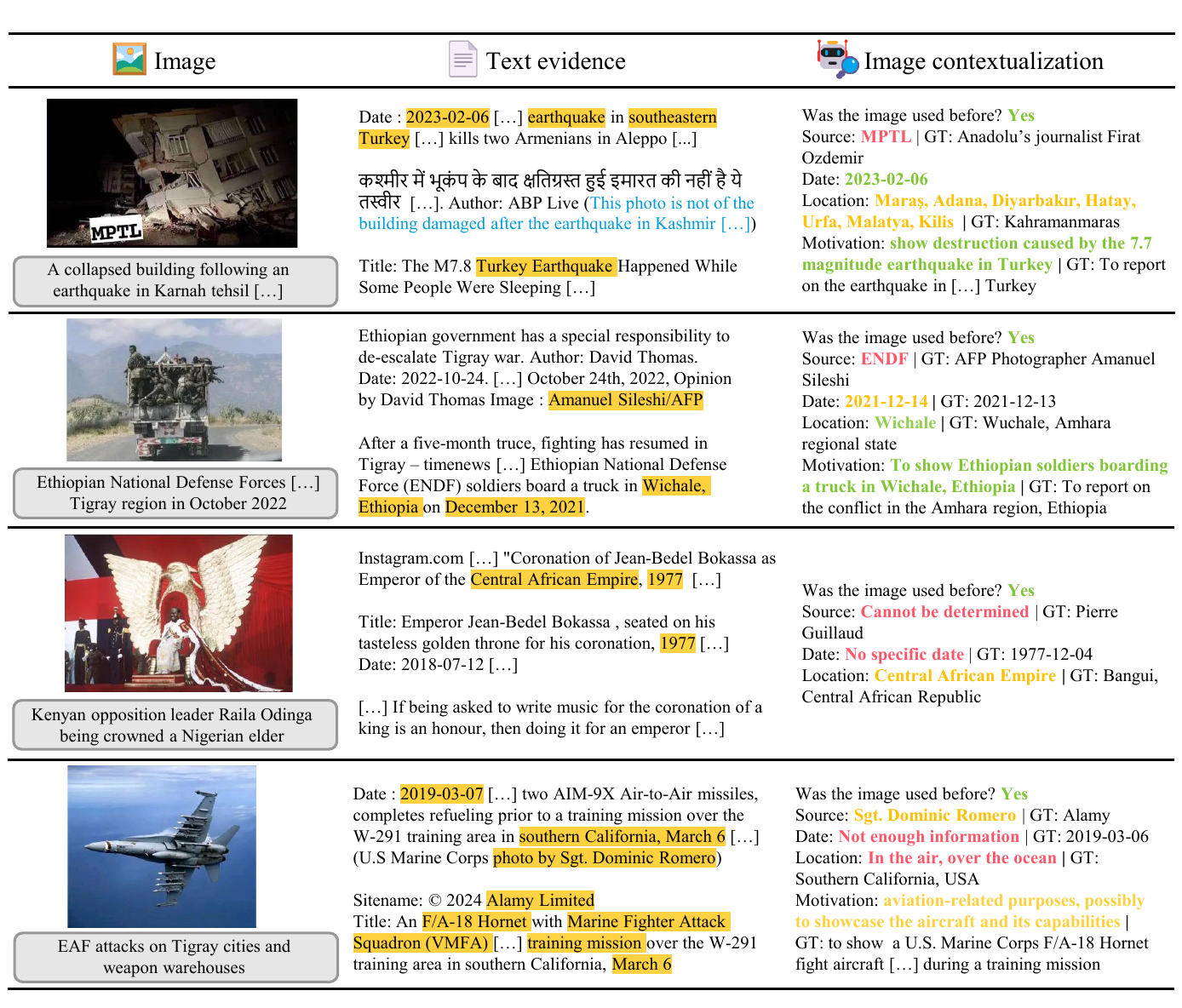}
    \caption{Images from the 5Pils test set with zero-shot multimodal GPT4 predictions. The middle column shows excerpts of the evidence texts with \hl{relevant snippets}. The right column shows \textcolor{mygreen}{\textbf{correct}}, \textcolor{myorange}{\textbf{partially correct}}, and \textcolor{myred}{\textbf{wrong}} predictions.  GT is the ground truth answer.}
    \label{fig:case-study}
\end{figure*}

\begin{figure}[!ht]
    \centering
    \includegraphics[scale=0.6]{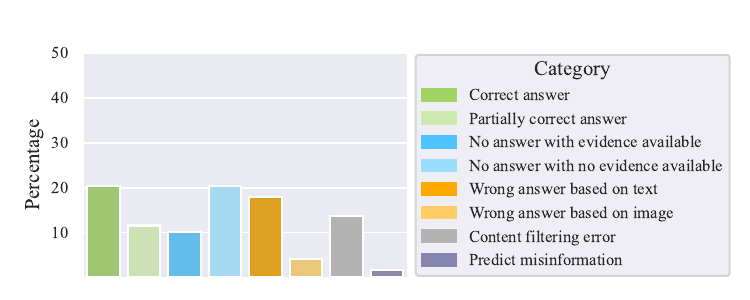}
    \caption{Qualitative analysis of zero-shot multimodal GPT4 answers (\%). }
    \label{fig:qualitative_total}
\end{figure}
 
\subsection{Qualitative analysis}

\begin{figure}
    \centering
    \includegraphics[scale=0.26]{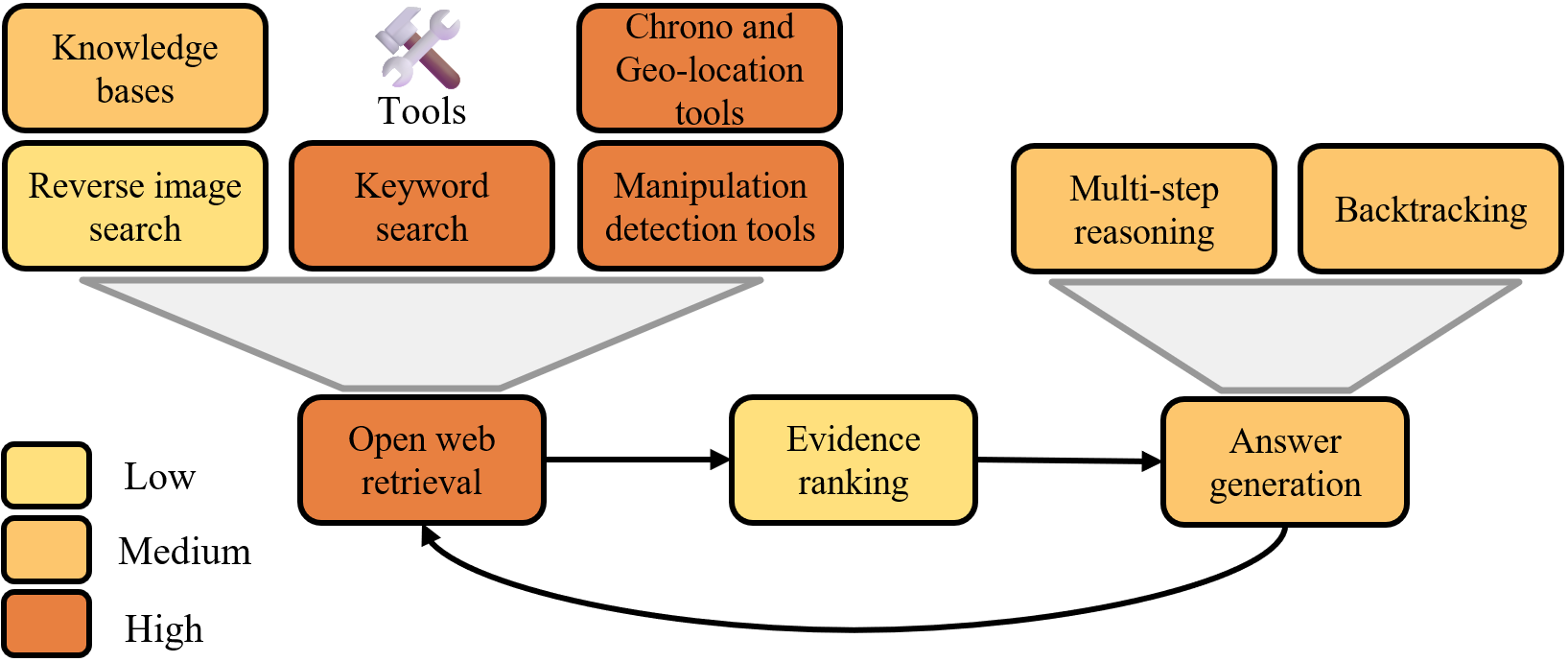}
    \caption{Open challenges in image contextualization, ranked by level of complexity.}
    \label{fig:challenges}
\end{figure}

We randomly sample 100 test images and manually categorize the predictions of multimodal zero-shot GPT4. Four examples are provided in Figure \ref{fig:case-study}. 31.9\% of the answers are (partially) correct.  We classify the other answers into six error categories.  The distribution of the correct and error categories is shown in Figure \ref{fig:qualitative_total}. We conduct the same analysis for zero-shot multimodal Llava and break down the correct and error categories distribution per pillar in Appendices \ref{sec:llava_quali} and \ref{sec:categories}, respectively.
20.4\% of the answers accurately state that not enough information is available in the image and text evidence, for example, in the \nth{3} image of Figure \ref{fig:case-study}, the Source cannot be identified. For 10.2\% of the answers, the response is the same, but sufficient information was actually available. This problem arises for the \nth{4} image, where useful evidence is directly available but unexploited. In the \nth{3} image, while the exact date is not given, the information on the event (``Coronation of Jean-Bedel Bokassa'') should be sufficient to predict the date based on world knowledge. For 17.9\% of the answers, the response is wrong and based on irrelevant evidence. Wrong answers based on wrongly interpreted image content are comparatively lower, at 4.2\%. 13.7\%  of the answers are not generated because of the GPT4 content filter. This mainly prevents GPT4 from contextualizing images of conflicts and violent incidents. Given the significant proportion of these images in FC articles, this constitutes a serious limitation to the use of GPT4. Lastly, 1.8\% of the answers contain misinformation elements.\\
From this analysis, we conclude that increasing the number of evidence retrieved from the open web constitutes the first direction to improve performance. The lack of evidence constitutes a regular challenge for human fact-checkers too. In the absence of RIS results, they turn to their world knowledge and keyword search to gather evidence, as illustrated in Appendix \ref{sec:examples_human_fc}. For the Date and Location pillars, this can be complemented by chrono- and geolocation tools.\\
Planning and combining the output of those tools constitute an interesting research direction for multimodal LLMs. As the number of evidence grows, the reasoning can be improved by breaking it down in several steps, interleaving evidence retrieval and reasoning, and implementing backtracking mechanisms \citep{NEURIPS2023_029df12a}. Figure \ref{fig:challenges} shows open challenges for the task, ranked by estimated level of complexity.

\section{Conclusion}
In this work, we fill an important gap in the field of visual misinformation by introducing the task of automated image contextualization. We create 5Pils, a dataset of 1,676 fact-checked images with meta-context QA pairs based on the 5 Pillars framework. We define a first baseline for the task, which combines the image with textual evidence retrieved from the open web. Despite achieving promising performance, several challenges remain. Directions for future work include better open web retrievers that combine multiple tools,  improved answer generation based on multi-step reasoning, and better methods for removing image manipulations.

\section{Limitations}

The 5Pils dataset has multiple limitations.\\
Firstly, we restricted data collection to articles written in English. While the 5Pils images are spread all over the world, this restriction results in the under-representation of certain regions, such as Latin America and Oceania. We analyze in Appendix \ref{sec:geographic_results} the impact of this under- and over-representation by region on performance. On the other hand, our dataset contains many images from Sub-Saharan Africa, a region underrepresented in previous misinformation datasets, compared to North America, South Asia, and China \citep{zlatkova-etal-2019-fact,10.1145/3539618.3591896,suryavardan2023factify}. In future work, 5Pils can scale to more regions and languages by collecting images and articles from other IFCN members.\\
Secondly, while we implemented a filter to remove evidence from FC articles, we cannot ensure a perfect recall. In the \nth{1} example of Figure \ref{fig:case-study}, one of the evidence is a FC article that passed our filter.\\
Thirdly, the dataset contains a majority of images that have effectively been used for misinformation purposes. This is a common limitation when relying on real-world FC articles \citep{zlatkova-etal-2019-fact,NEURIPS2023_cd86a305}, as human fact-checkers invest more time in writing articles about false claims than true claims. Nevertheless, our analysis in Appendix \ref{sec:source_analysis} reveals that the images of 5Pils have very diverse origins, ranging from social media users to local and global news agencies.\\ 
Lastly, the metrics for Source are slightly pessimistic with regard to the model's performance. If the prediction is more specific than the ground truth, no or partial credit is given. On the \nth{4} example of Figure \ref{fig:case-study}, GPT4 accurately predicts the photographer's name, while the FC article only reports the website where the image can be found. This results in a score of 0, while the prediction is actually better than the ground truth.  \\
Regarding our baseline, we find two limitations.\\
Firstly, not all RIS results can be scraped from the open web. Evidence from social media platforms is typically not accessible. In some cases, image contextualization requires access to such social media posts. This constitutes an important difference with human fact-checkers, who can access social media posts from their browser.\\
Secondly, our baseline does not assess the reliability of the retrieved evidence. Hence, there is no guarantee that the information they provide is accurate, and different evidence might contradict each other. Assessing reliability and using it to filter and rank text evidence constitutes an interesting direction for future work.

\section{Ethics statement}

Automated image contextualization is designed as a new tool to help fact-checkers verify the vast quantities of visual misinformation circulating on the internet. Automated approaches should respect the same code of practice as professional fact-checkers, which includes respect for online privacy. In particular, for the Source pillar, techniques for automated profiling on social media should not be used for automated image contextualization. Care should also be taken when publishing the results, as they could put the Source at risk.\\
The 5Pils dataset was collected from publicly available FC organization websites. Due to the real-world nature of the data, events covered include wars and conflicts. As a result, some images contain graphic, violent content. When collecting the data, we decided not to filter out images with violent content to cover the actual distribution of images that our target users, professional fact-checkers, would want to provide as input. Given the graphic nature of some images, we do not release them directly. Instead, we do publicly release the URLs of the FC articles and extracted evidence, as well as the script that allows to collect and process the images and the evidence text.

\section*{Acknowledgements}

This work has been funded by the LOEWE initiative (Hesse, Germany) within the emergenCITY center (Grant Number: LOEWE/1/12/519/03/05.001(0016)/72) and by the German Federal Ministry of Education and Research and the Hessian Ministry of Higher Education, Research, Science and the Arts within their joint support of the National Research Center for Applied Cybersecurity ATHENE. We gratefully acknowledge the support of Microsoft with a grant for access to OpenAI GPT models via the Azure cloud (Accelerate Foundation Model Academic Research).
We want to express our gratitude to Max Glockner, Haritz Puerto, Jiahui Geng, Luke Bates, Sukannya Purkayastha, and Aniket Pramanick for their insightful comments on a draft of this paper.

\bibliography{anthology,anthology_p2,custom}
\bibliographystyle{acl_natbib}

\appendix

\section{Analysis of the FC narrative}
\label{sec:epistemic}

\begin{figure}
    \centering
    \includegraphics[scale=0.17]{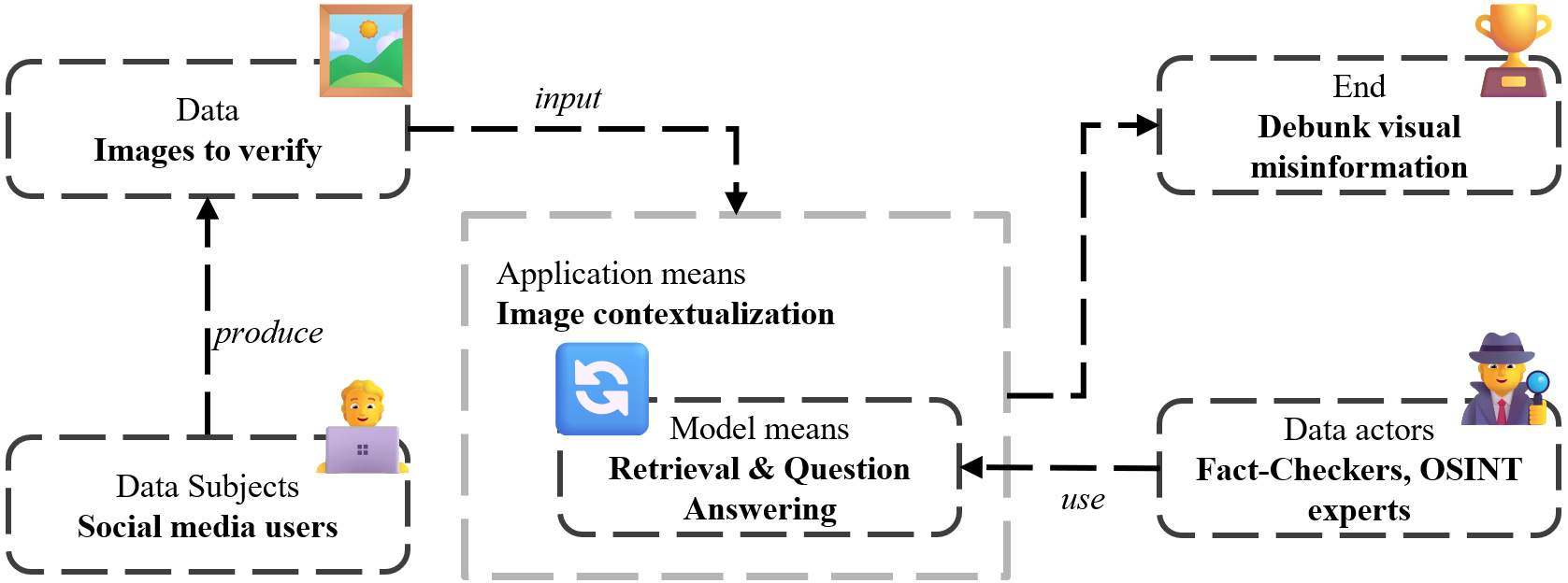}
    \caption{Image contextualization narrative diagram. Figure based on \citet{schlichtkrull-etal-2023-intended}.}
    \label{fig:intended-uses}
\end{figure}

\begin{figure}
    \centering
    \includegraphics[scale=0.27]{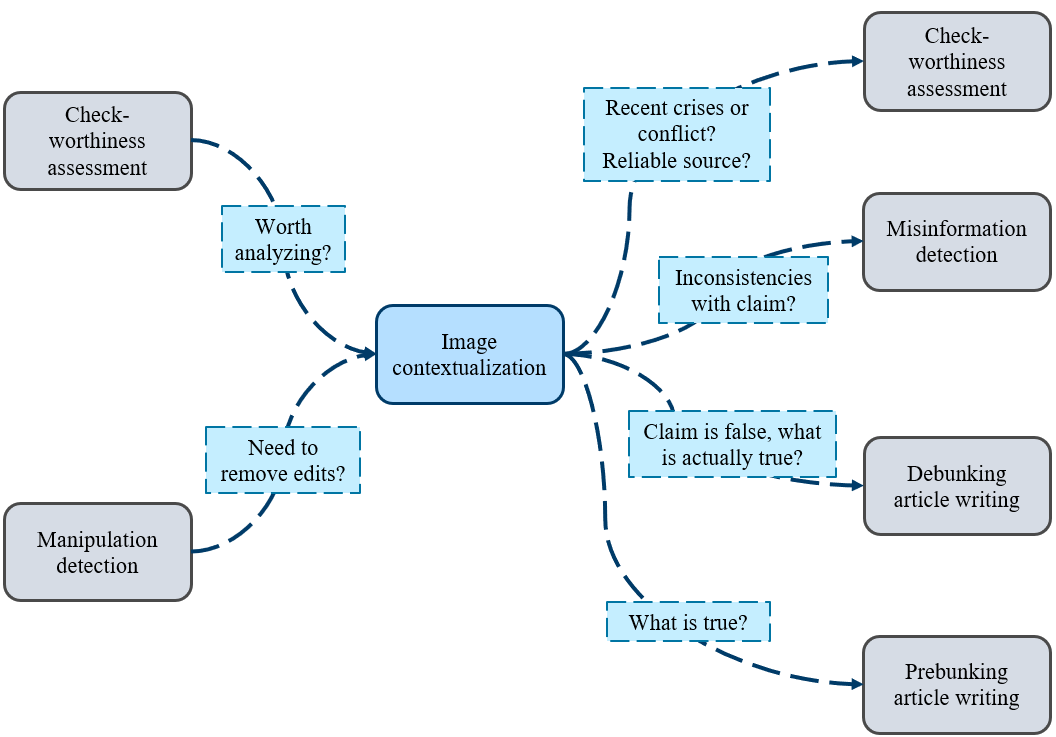}
    \caption{Connections of image contextualization with other components of the human fact-checking pipeline.}
    \label{fig:fc_pipeline}
\end{figure}

Following the recommendations of \citet{schlichtkrull-etal-2023-intended}, we analyze the FC narrative introduced in this paper. Figure \ref{fig:intended-uses} provides a diagram of the different epistemic elements of our narrative. We envision fact-checkers as the primary data actors that will use our automated retrieval and QA pipeline (model means) to predict the original meta-context of images (application means) and debunk visual misinformation (ends). Other potential data actors include experts in Open Source Intelligence (OSINT). The data subjects are primarily social media users who will share image content that might be out-of-context, manipulated, or fake.\\
Figure \ref{fig:fc_pipeline} shows the interactions of image contextualization with other components of the human FC pipeline. We identify two components at the input of image contextualization. (1) Assessing check-worthiness saves computational resources by filtering out content that is unlikely to be visual misinformation. (2) Detecting manipulations informs whether any edits need to be removed to obtain the original image. For the output, we identify four components covered in Section \ref{sec:intro}. In this work, we integrate manipulation detection in our automated baseline, while other tasks at the input and output are expected to be performed by human fact-checkers.  Integrating image contextualization with other automated FC tasks is an interesting direction for future work.

\section{Image contextualization with keyword search}
\label{sec:examples_human_fc}

\begin{figure}
    \centering
    \includegraphics[scale=0.3]{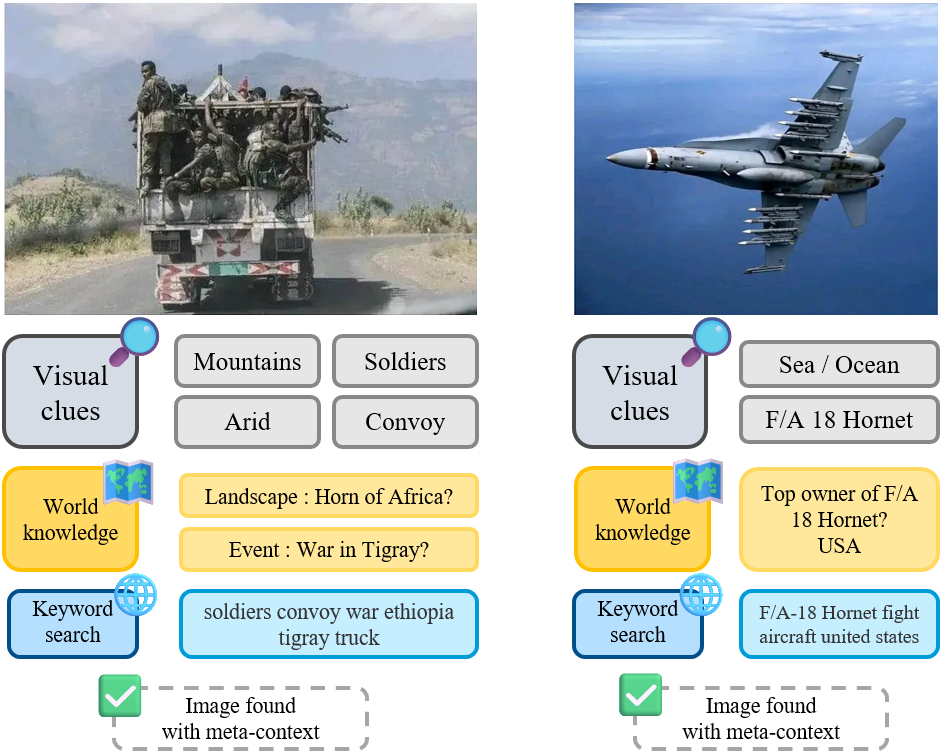}
    \caption{Contextualization with keyword search.}
    \label{fig:examples_no_ris}
\end{figure}

It can happen that RIS engines do not return any results for a given image. In those cases, human fact-checkers use other methods to retrieve evidence, including keyword search. We provide two examples in Figure \ref{fig:examples_no_ris}, corresponding to the \nth{2} and \nth{4} images of Figure \ref{fig:case-study}.\\
\textbf{Left image}: The image contains visual cues, namely, a military convoy, a semi-arid landscape, and mountain chains in the background. Assuming sufficient world knowledge, we know that this landscape is characteristic of the Horn of Africa, and the main recent conflict in the area was the war in Tigray. Hence, we make the hypothesis that the image is related to the war in Tigray. We run the web search query “soldiers convoy war ethiopia tigray truck”. The 10th result is the same image hosted on the platform Getty, captioned with all relevant meta-context. \\
\textbf{Right image}: the plane is recognizable as a Boeing F/A-18 Hornet fighter aircraft. Searching Wikipedia for information, we find that the top owner of F/A-18 is the US Airforce. We run the following web search query: “F/A-18 Hornet fight aircraft united states”, and retrieve the target image among the first results on a web-page that contains all the meta-context information. \\

\section{Contextualization of manipulated and fake images}
\label{sec:examples}

\begin{figure}
    \centering
 \includegraphics[scale=0.45]{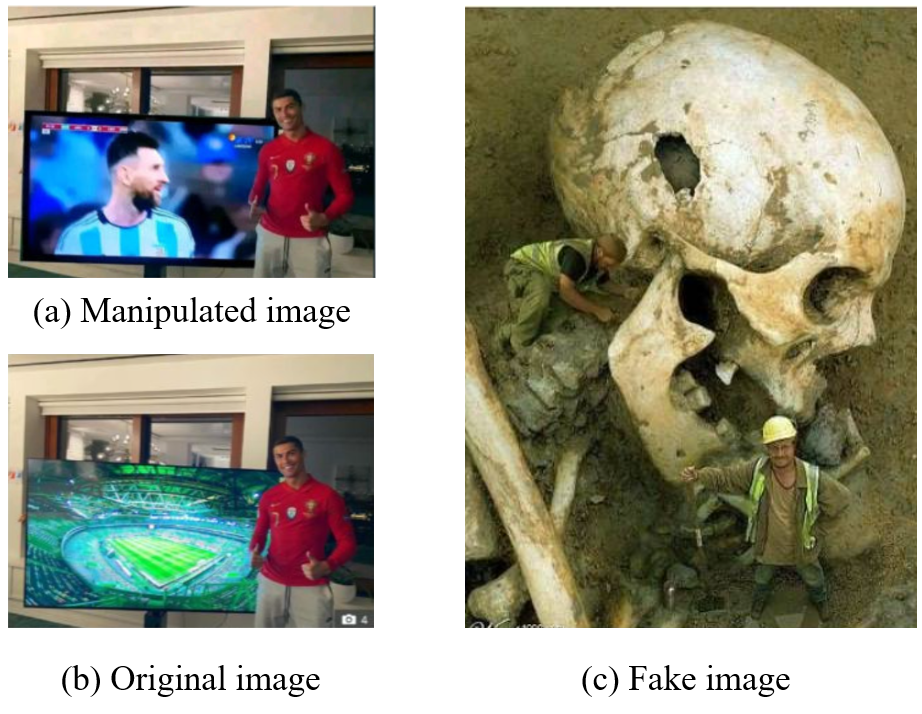}
    \caption{A manipulated image with edited television background (a) and the original version (b). A digital artwork of a giant skeleton (c).}
    \label{fig:manipulated-fake}
\end{figure}

We provide examples of manipulated and fake images in Figure \ref{fig:manipulated-fake}. The manipulated image (a) claimed that footballer Cristiano Ronaldo was showing support for the Argentine football team.\footnote{\href{https://factly.in/this-post-shares-an-edited-image-to-claim-that-cristiano-ronaldo-is-supporting-argentina/}{factly.in/this-post-shares-an-edited-image-to-claim-that-cristiano-ronaldo-is-supporting-argentina}} By collecting RIS results, it was first observed that the image had been manipulated (b). Additional meta-context elements informed the fact-checkers that the image was taken prior in time, when the footballer showed support for the Portugal team and originated from his official social media account. Image (c) claimed to show the discovery of giant human skeletons. \footnote{\href{https://factly.in/a-digitally-created-artwork-falsely-shared-as-a-real-picture-of-a-giant-human-skeleton/}{factly.in/a-digitally-created-artwork-falsely-shared-as-a-real-picture-of-a-giant-human-skeleton}} By searching the Source and Motivation pillars, it was discovered that an artist created the image for a digital artwork competition.

\section{Automated annotations prompt}
\label{sec:prompt}

\begin{figure}
    \centering
    \includegraphics[scale=0.6]{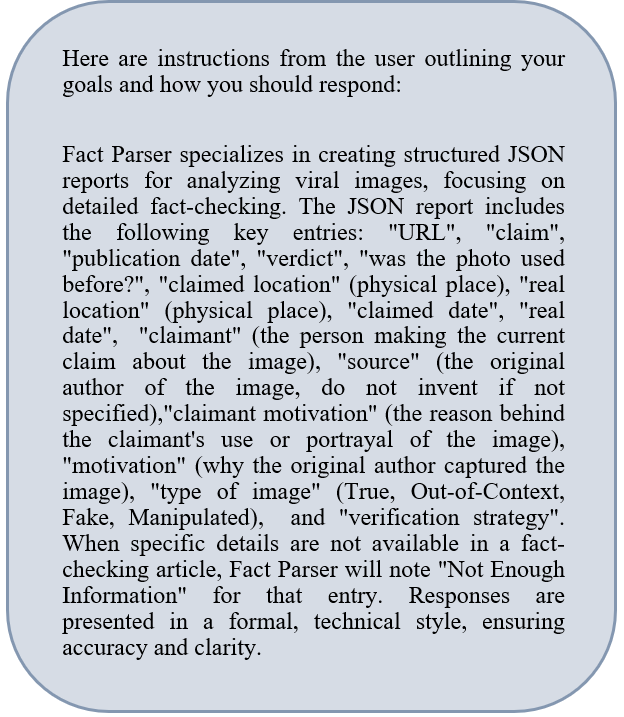}
    \caption{Prompt for GPT4 annotations of FC articles.}
    \label{fig:labeling_prompt}
\end{figure}

\begin{figure}
    \centering
    \includegraphics[scale=0.45]{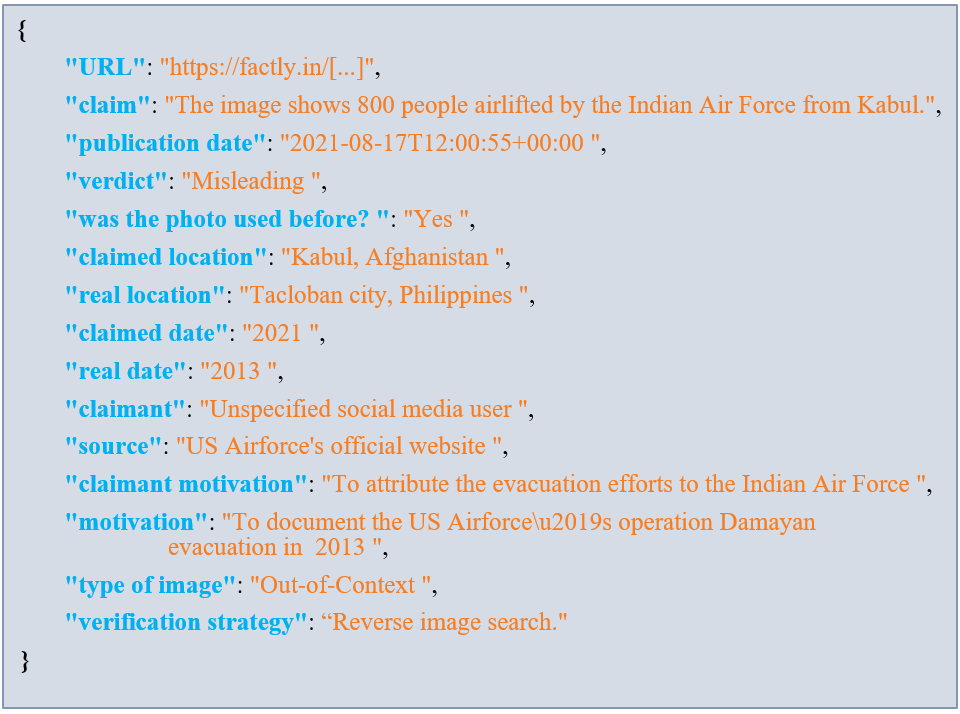}
    \caption{Example of GPT4 annotations.}
    \label{fig:example_labeling}
\end{figure}

Figure \ref{fig:labeling_prompt} shows the instructions given to GPT4 to extract the 5 Pillars answers from FC articles. The prompt was improved iteratively over batches of 10 articles until a satisfactory performance was achieved. The prompt instructs GPT4 to create a JSON file containing various fields. If the information is unavailable, the standard answer should be  ``Not enough information''. Figure \ref{fig:example_labeling} reports the output for the image of Figure \ref{fig:intro}. The \textit{claimed location}, \textit{claimed date}, \textit{claimant}, and \textit{claimant motivation} fields are requested to make sure that GPT4 clearly distinguishes the claimed context of the image from its original meta-context. The \textit{verification strategy} was initially extracted by querying GPT4. Later, it was replaced by a keyword matching over the original article. GPT4 sometimes extracts a list of strings, for example, when an image is composite. In those cases, we merge the list items into a string. The data labeling was performed using the Azure OpenAI service with API version \textit{2023-10-01-preview}. The content filter was turned off to allow GPT4 to label FC articles that discuss conflicts and other events that involve violence.

\section{Co-occurrence of 5 pillar answers}
\label{sec:cooccurence}

\begin{figure}[!ht]
    \centering
    \includegraphics[scale=0.35]{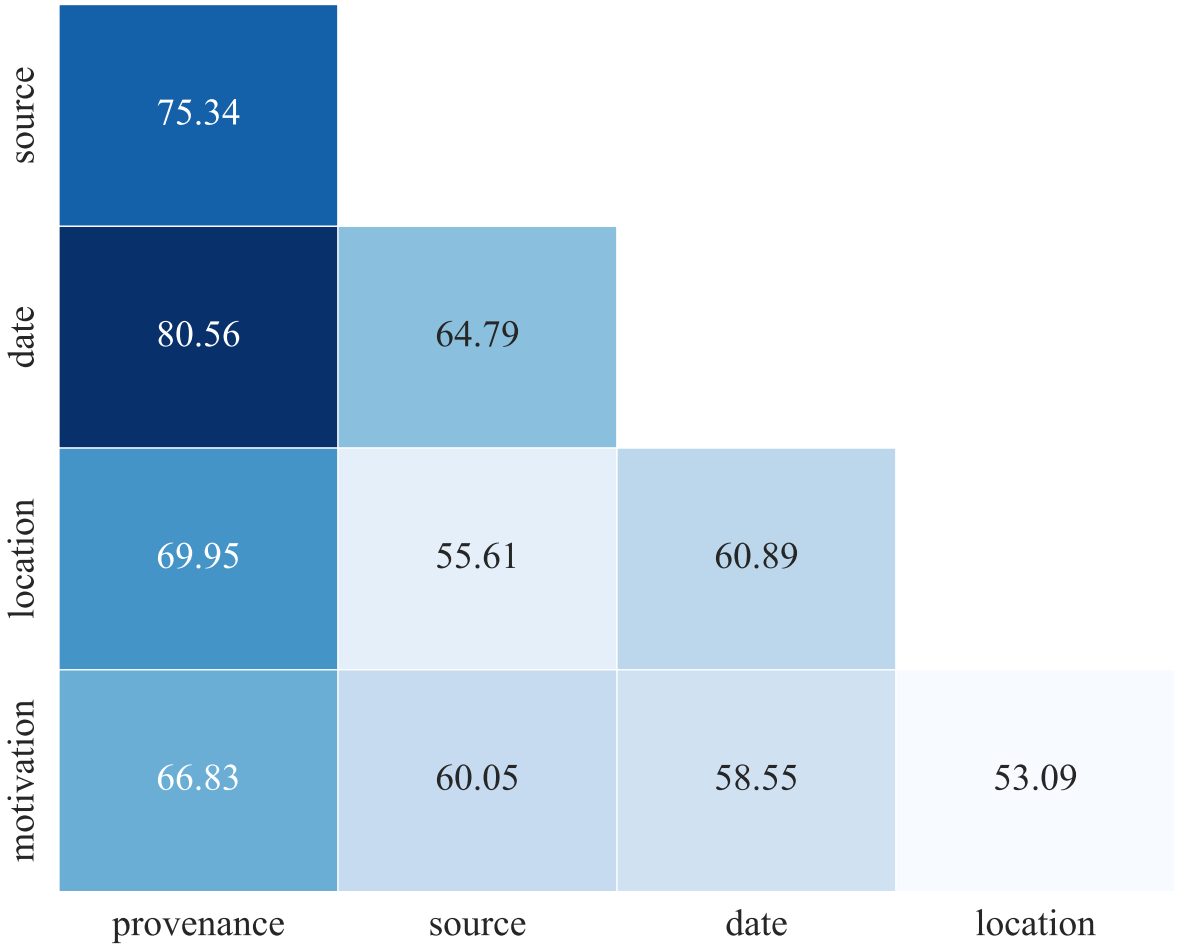}
    \caption{Co-occurrence of 5 Pillars answers in the dataset (\%).}
    \label{fig:cm}
\end{figure}

Figure \ref{fig:cm} shows the percentage of images for which two pillars are answered together. The percentages are coherent with the total occurrence of each pillar in the dataset. For example, the Date-Source is higher than the Date-Location co-occurrence because there are more images with Source answers than images with Location answers. No specific over- or under-representation of a pair is observed.

\section{Type of misinformation over time}
\label{sec:misinfo_type}
\begin{figure}
    \centering
    \includegraphics[scale=0.25]{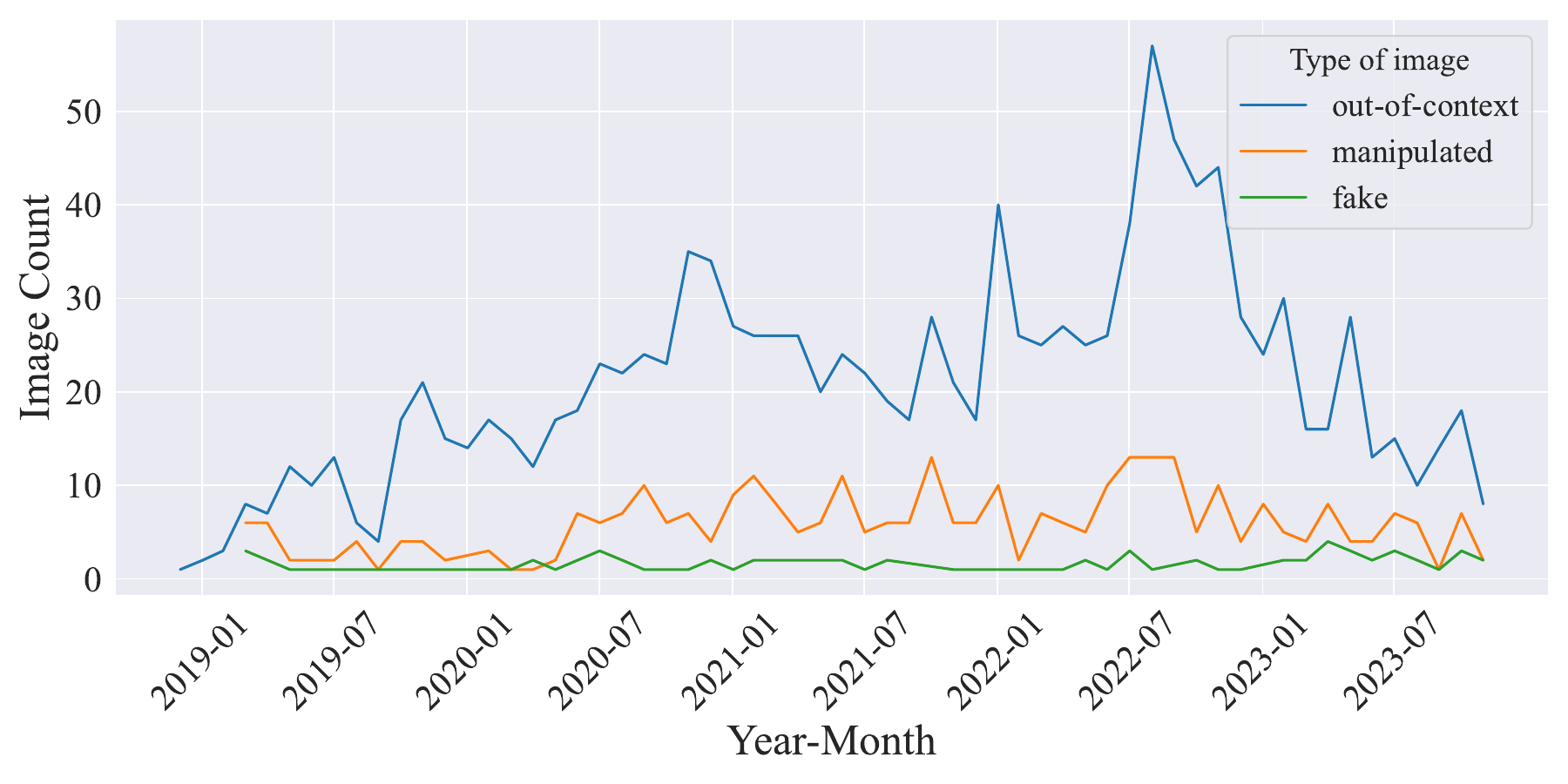}
    \caption{ Time evolution of the type of misinformation in the dataset.}
    \label{fig:time-analysis}
\end{figure}

Figure \ref{fig:time-analysis} displays the type of images verified by the FC organizations over time. Despite the rise of generative AI technologies, Fake images still represent a stable and low share of the verified content. Out-of-context consistently remains the most popular type of visual misinformation. The peak in out-of-context images in the summer of 2022 corresponds to the Kenyan general elections and the end of the conflict in Tigray.

\begin{figure*}
    \centering
    \includegraphics[scale=0.63]{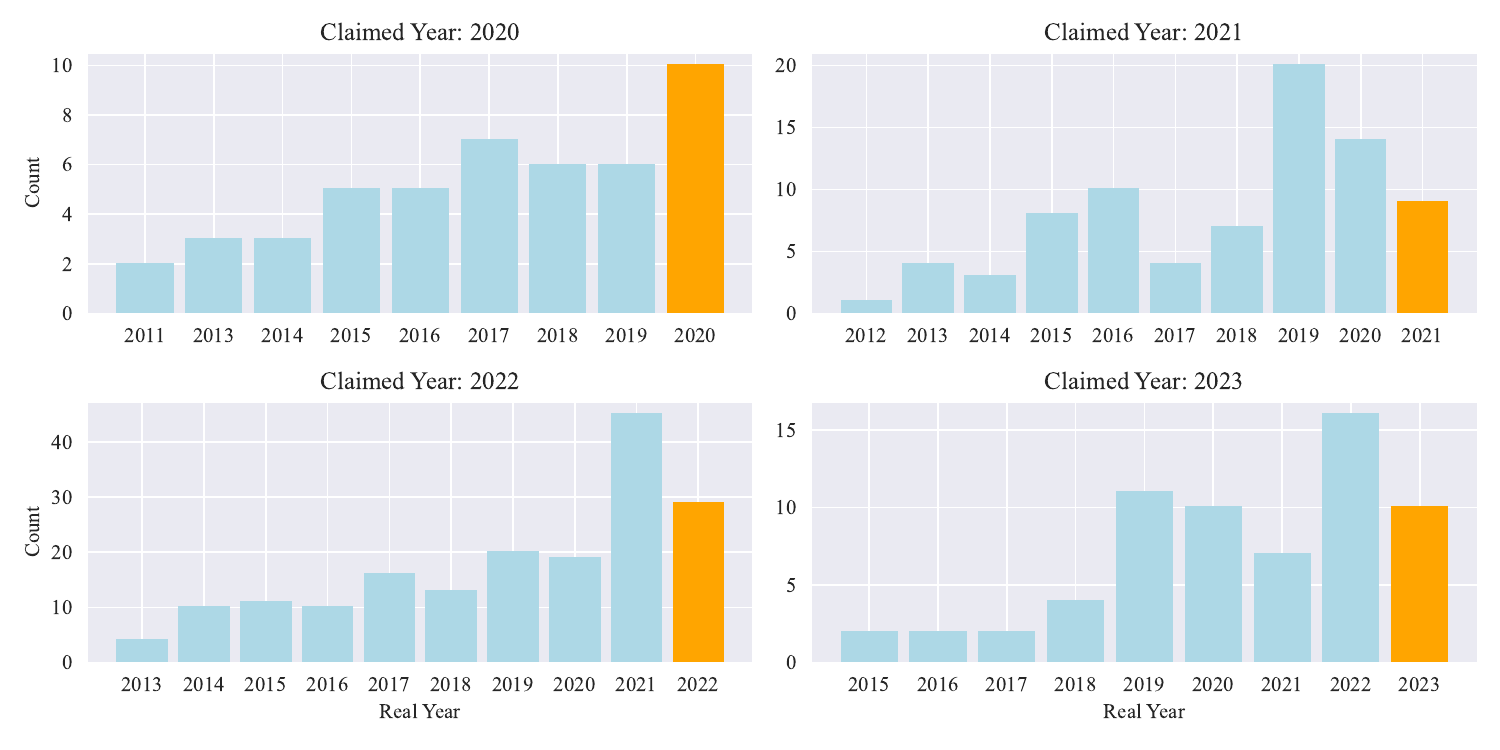}
    \caption{The distribution of real dates over time for images with a given claimed date, in years. The orange bar indicates images for which the claimed date matches the real date.}
    \label{fig:temporal_analysis}
\end{figure*}

\section{Source analysis}
\label{sec:source_analysis}

We investigate the distribution of the Source pillar across platforms and organizations. We group the sources into main categories using regular expressions. Table \ref{tab:source} reports the 10 most frequent sources in 5Pils. At least 12.95\% of the images were first published by social media users. Interestingly, more than 10\% of the images originally come from news agencies, more than 5\% from image-sharing platforms, and many more from various local newspapers and private web pages. This diversity tends to reveal that many, if not any, images related to news events can be repurposed for misinformation.

\section{Temporal analysis}
\label{sec:temporal}

\begin{table}
    \centering
    \begin{tabular}{cc}
    \hline
     Source    & Frequency  \\
    \hline
      Twitter user  & 5.73 \\
      Facebook user  & 5.73 \\
      Getty Images & 5.57 \\
      Reuters & 4.55 \\
      AP & 2.51 \\
      BBC & 1.57\\
      Instagram & 1.49 \\
      AFP & 1.18 \\
      ANI & 1.10 \\
      EPA & 0.71 \\
    \hline
    \end{tabular}
    \caption{Most frequent sources in 5Pils (\%).}
    \label{tab:source}
\end{table}

\begin{figure*}
    \centering
    \includegraphics[scale=0.42]{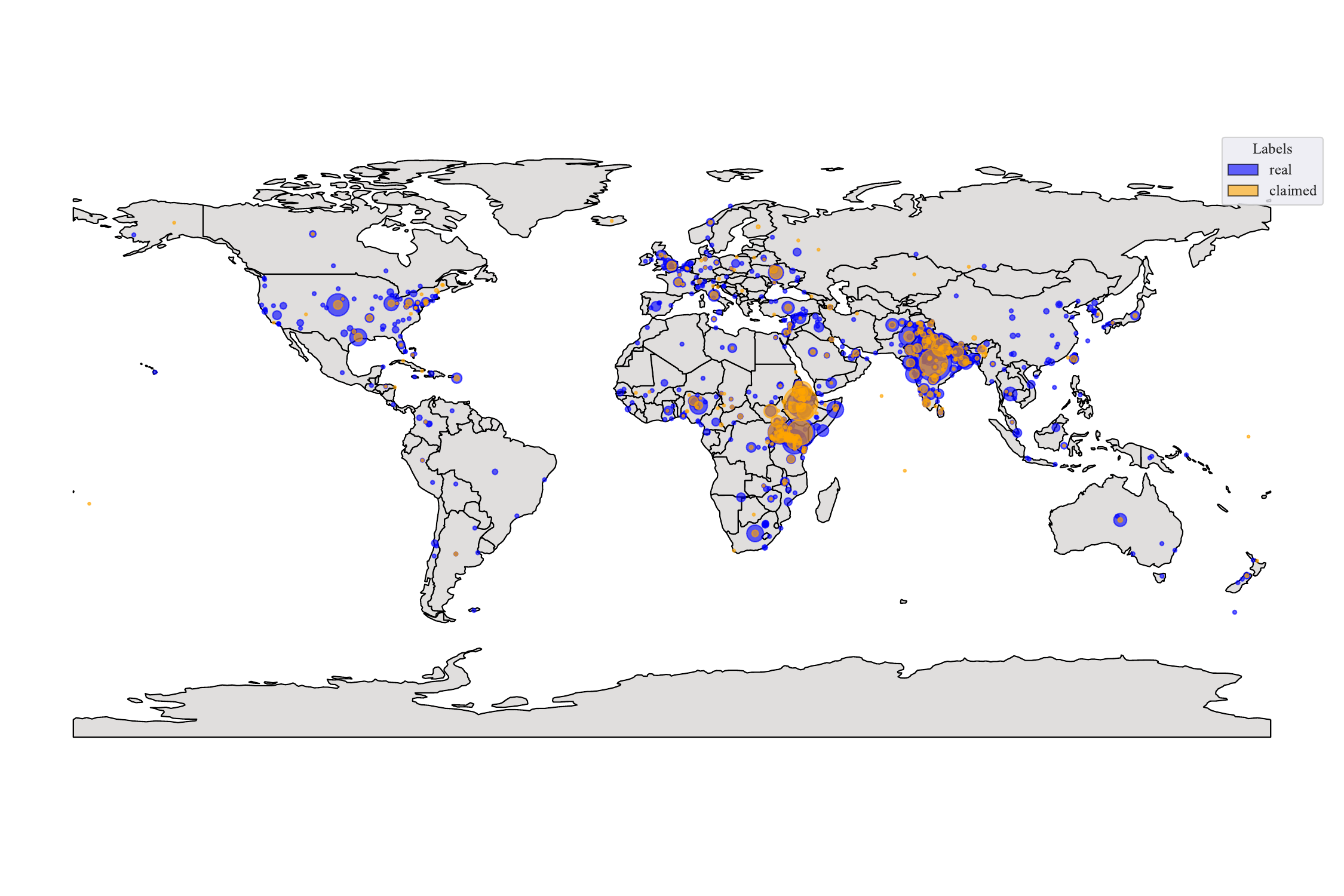}
    \caption{\textcolor{orange}{\textbf{Claimed}} and \textcolor{blue}{\textbf{real}} locations of 5Pils images. The size of the circle is proportional to the number of images with these coordinates.}
    \label{fig:world-map}
\end{figure*}

\begin{figure*}
    \centering
    \includegraphics[scale=0.42]{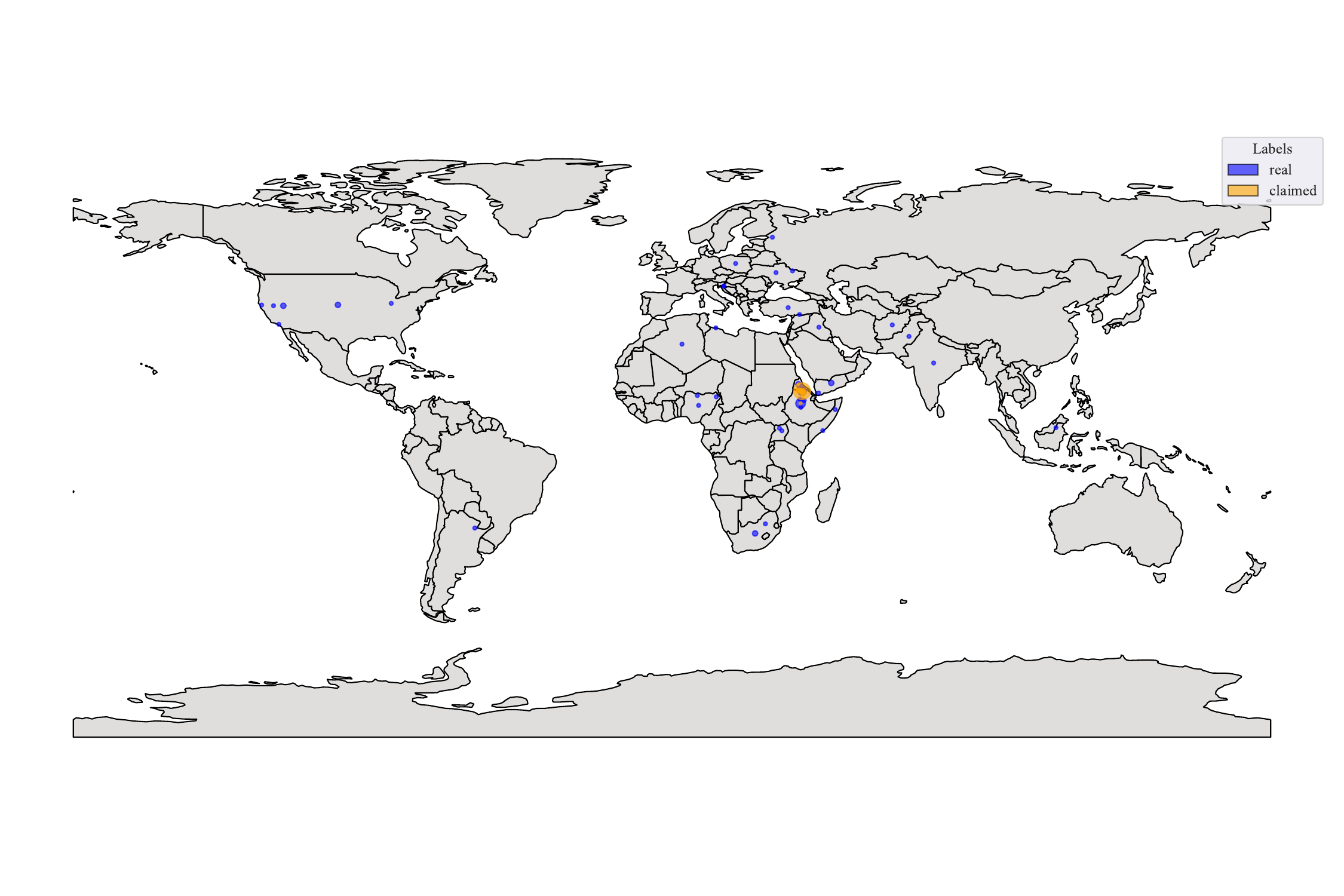}
    \caption{\textcolor{orange}{\textbf{Claimed}} and \textcolor{blue}{\textbf{real}} locations of images with the conflict in Tigray as claimed context. The size of the circle is proportional to the number of images with these  coordinates.}
    \label{fig:tigray}
\end{figure*}

\begin{table*}[!ht]
\centering
\resizebox{\textwidth}{!}{ %
\begin{tabular}{ccccccccccccccc}
\hline
\textbf{} &   \multicolumn{2}{c}{\textbf{Source}} & & \multicolumn{2}{c}{\textbf{Date}} & & \multicolumn{4}{c}{\textbf{Location}} &   & \multicolumn{3}{c}{\textbf{Motivation}} \\
\textbf{$k$}  &  \textbf{RougeL} & \textbf{Meteor} & & \textbf{EM} & \textbf{$\Delta$} & & \textbf{RougeL} & \textbf{Meteor} &  \textbf{HL$\Delta$} &   \textbf{CO$\Delta$}  & & \textbf{RougeL} & \textbf{Meteor}  & \textbf{BertS} \\ 
 \hline
1 & \textbf{6.5}$_{\pm0.2}$  & \textbf{4.2}$_{\pm0.2}$ & & 2.1$_{\pm0.6}$ & 42.7$_{\pm0.2}$ & & 27.3$_{\pm0.2}$ & 20.0$_{\pm0.2}$ &  33.0$_{\pm0.6}$ &  39.9$_{\pm0.4}$ & & 6.2$_{\pm0.4}$  & 3.2$_{\pm0.4}$  & 68.7$_{\pm0.2}$\\
3 & 4.5$_{\pm0.2}$  & 2.4$_{\pm0.2}$  & & 4.1$_{\pm0.3}$ &  \textbf{42.9}$_{\pm0.7}$ & & 28.7$_{\pm0.3}$ & 20.9$_{\pm0.3}$ &  33.7$_{\pm0.4}$ &  40.5$_{\pm0.5}$ & & 6.7$_{\pm0.3}$ & 3.4$_{\pm0.3}$ & 68.4$_{\pm0.1}$\\
5  & 5.3$_{\pm0.2}$ & 3.3$_{\pm0.2}$ & & 2.4$_{\pm0.3}$ & 42.7$_{\pm0.2}$ & & \textbf{28.9}$_{\pm0.9}$ & \textbf{21.6}$_{\pm0.9}$ &  \textbf{35.6}$_{\pm0.3}$ &  \textbf{42.6}$_{\pm0.2}$ & & 6.6$_{\pm0.6}$  &  3.3$_{\pm0.6}$ & 68.4$_{\pm0.5}$ \\
10 & 5.9$_{\pm0.5}$ & 3.9$_{\pm0.5}$ & & \textbf{4.6}$_{\pm0.3}$ & 42.1$_{\pm0.7}$ & & 26.1$_{\pm0.4}$ & 19.6$_{\pm0.4}$&  31.6$_{\pm0.3}$ & 38.8$_{\pm0.3}$ & & \textbf{7.7}$_{\pm0.0}$  &  \textbf{3.7}$_{\pm0.0}$ & \textbf{69.5}$_{\pm0.0}$\\
\hline

\end{tabular}}

\caption{Multimodal zero-shot Llava results on the validation set for different values of $k$, the number of text evidence provided in the prompt (\%). Reported results are averages over 3 iterations with standard deviations. The best scores are marked in \textbf{bold}.}
\label{tab:results_hyperparam}
\end{table*}

We study the distribution of the real image dates for a given claimed date. We report the distributions for images claimed to be from 2020, 2021, 2022, and 2023 in Figure \ref{fig:temporal_analysis}. Most real dates are concentrated in the 3 to 4 years that include and precede the claimed date. A downward trend is then observed going back in time. This tells us that visual misinformation is often based on recent images. This information could be used as a heuristic to guide the prediction of the Date pillar.

\section{Spatial analysis}
\label{sec:geographic}
We look deeper into the Location pillar and investigate whether there are any noticeable geographic patterns in the way images from one location are used in a claim about another location. Using Nominatim,\footnote{\href{https://geopy.readthedocs.io/en/stable/}{geopy.readthedocs.io}} we obtain coordinates for most of the claimed and real locations of the images in the dataset.  Figures \ref{fig:world-map} show the spatial distribution of all images that could be mapped to Nominatim coordinates. Large dots at the geographic center of a country indicate that the location is the country itself. The dataset covers a large diversity of real locations. East Africa and South Asia are predominant. This is expected because the FC organizations from which we collected the data are located in India, Kenya, and South Sudan. Despite that, Europe, North America, and other parts of Africa get important coverage, too. Interestingly some countries have few or no claims about them, but their images are nevertheless frequently used out-of-context. This includes China, Syria, Iraq, and Spain.  
In a second analysis, we study the dispersion of real locations. To do this, we looked at one specific world event where the claimed locations are concentrated in a limited geographic area. We plotted all verified images claiming to be associated with the war in Tigray, a conflict between 2020 and 2022 in the Tigray region of Ethiopia, close to Eritrea. The resulting map is shown in Figure \ref{fig:tigray}. While the claimed locations are concentrated in Northern Ethiopia and Eritrea, as expected, only a part of the real locations are located there, too, while the rest is spread out over the globe. Images can be found as far away as the United States. Those are mainly images of drones and aircraft training exercises. This large geographical dispersion illustrates the challenge and importance of predicting the Location pillar.

\section{Few-shot prompt template}
\label{sec:prompts}

\begin{figure}
    \centering
    \includegraphics[scale=0.42]{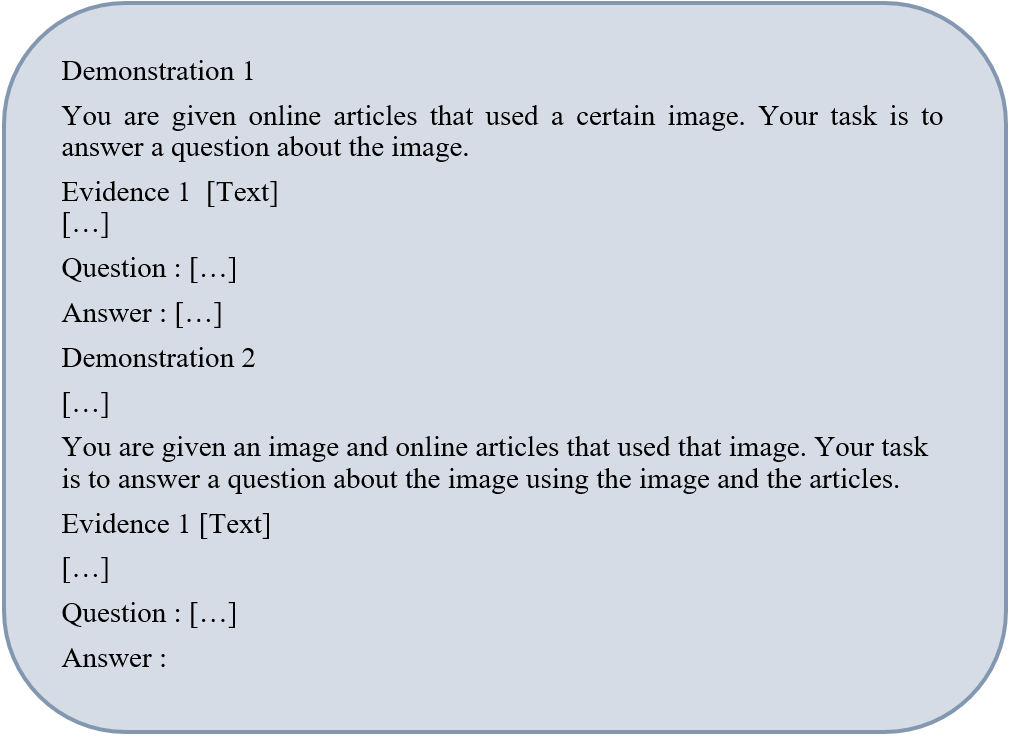}
    \caption{Few-shot prompt template for GPT4 with multimodal input.}
    \label{fig:template}
\end{figure}

Figure \ref{fig:template} shows the few-shot prompt template provided to GPT4 for multimodal answer generation. Demonstration and evidence are numbered. The zero-shot prompt follows the same structure, except that the demonstrations are removed. Different instructions are provided when the input modality is the image (``You are given an image. Your task is to answer a question about the image.'') or the text evidence (``You are given online articles that used a certain image. Your task is to answer a question about the image.''). Adjustments are made to the prompts for Llama2 and Llava to include the special tokens expected by those  LLMs. For Llava, every prompt starts with ``USER: '' and ``Answer: '' at the end of the GPT4 prompt is replaced by ``ASSISTANT: ''. For Llama2, every prompt starts with ``<s>[INST] <<SYS>> You are given [...] Your task is to answer a question about the image. <</SYS>>'', and ends with ``[/INST]''.

\section{Implementation details}
\label{sec:details}

\textbf{Manipulation detection} The ViT model is fine-tuned for 10 epochs on the train set with a learning rate of 2e-4.\footnote{\href{https://huggingface.co/google/vit-base-patch16-224}{huggingface.co/google/vit-base-patch16-224}} It achieves an F1 score of 51\% on the validation set, and 40\% on the test set.\\
\textbf{Evidence retrieval} We set the maximum number of URLs to return by the Google RIS engine to 50. We scrape the following fields of the web-pages using Trafilatura: the \textit{title}, \textit{description}, \textit{author}, \textit{hostname}, \textit{sitename}, and \textit{publication date}. We also retrieve the \textit{image} and its \textit{caption} if available using a custom script. In total, we retrieved and scraped 2284 web-pages for the test set. 64.1\% of the evidence is written in English.\\
\textbf{Evidence ranking} Embeddings are computed with a pre-trained multilingual CLIP model \citep{pmlr-v139-radford21a,carlsson-etal-2022-cross}.\footnote{\href{https://huggingface.co/M-CLIP/XLM-Roberta-Large-Vit-L-14}{huggingface.co/M-CLIP/XLM-Roberta-Large-Vit-L-14}} Due to budget constraints regarding the use of GPT4, we fixed the maximum number of evidence to use in the prompt $k$ to 3. We report in Table \ref{tab:results_hyperparam} the results obtained on the validation set for different values of $k$ with zero-shot multimodal Llava. The results suggest that increasing $k$ does not always improve the performance, as shown by the Location scores that decline when $k$ increases from 5 to 10. We conclude that better performance can be obtained by optimizing $k$ for each pillar separately.\\
\textbf{Answer generation} Demonstrations are selected with a nearest neighbor strategy \citep{liu-etal-2022-makes}. Given an image from the test set, we compute the similarity with images in the train set using CLIP and select the N nearest neighbors as demonstrations. For all LLMs, we set the temperature to 0.2 to allow some diversity while remaining close to a deterministic output. Experiments with Llama2 (\textit{meta-llama/Llama-2-7b-chat-hf}) and Llava (\textit{llava-hf/llava-1.5-7b-hf}) are run on a single A100 GPU. For  GPT4 (\textit{gpt-4-1106-vision-preview}), we conduct our experiments through the Azure OpenAI service with API version \textit{2023-10-01-preview}. The following questions are used as part of the prompts: ``Who is the source/author of the image? Answer with one or more person or entities in a few words.'' for Source, ``When was the image taken? Answer with one or more dates in a few words.'' for Date, ``Where was the image taken? Answer with one or more locations in a few words.'' for Location, ``Why was the image taken? Answer in a few words.'' for Motivation.\\

\section{Qualitative analysis of multimodal zero-shot Llava}
\label{sec:llava_quali}

\begin{figure}
    \centering
    \includegraphics[scale=0.6]{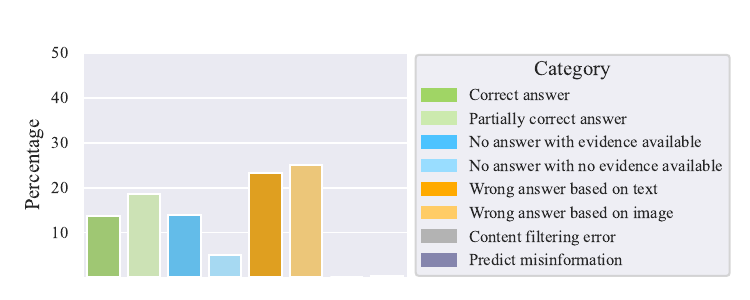}
    \caption{Qualitative analysis of zero-shot multimodal Llava answers (\%).}
    \label{fig:qualitative_analysis_llava_total}
\end{figure}

Figure \ref{fig:qualitative_analysis_llava_total} reports the results of the qualitative analysis of multimodal zero-shot Llava for a random sample of 100 test images. Llava makes less correct predictions than GPT4, 13.6\% against 20.4\%, but more partially correct predictions, 18.6\% against 11.6\%. There are two major differences with GPT4. Firstly, there is no content filter in place. Secondly, Llava does not often abstain from answering. Hence, more errors are due to wrong answers based on text and image content than with GPT4.

\section{Qualitative analysis per pillar}
\label{sec:categories}

\begin{figure*}
    \centering
    \includegraphics[scale=0.34]{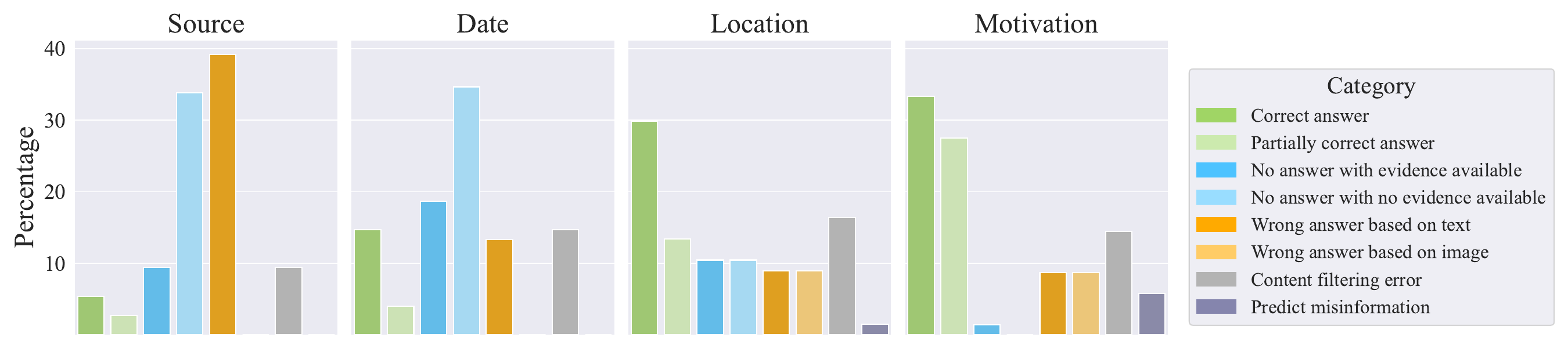}
    \caption{Qualitative analysis per pillar of multimodal zero-shot GPT4 answers (\%).}
    \label{fig:qualitative_analysis}
\end{figure*}

\begin{figure*}
    \centering
    \includegraphics[scale=0.34]{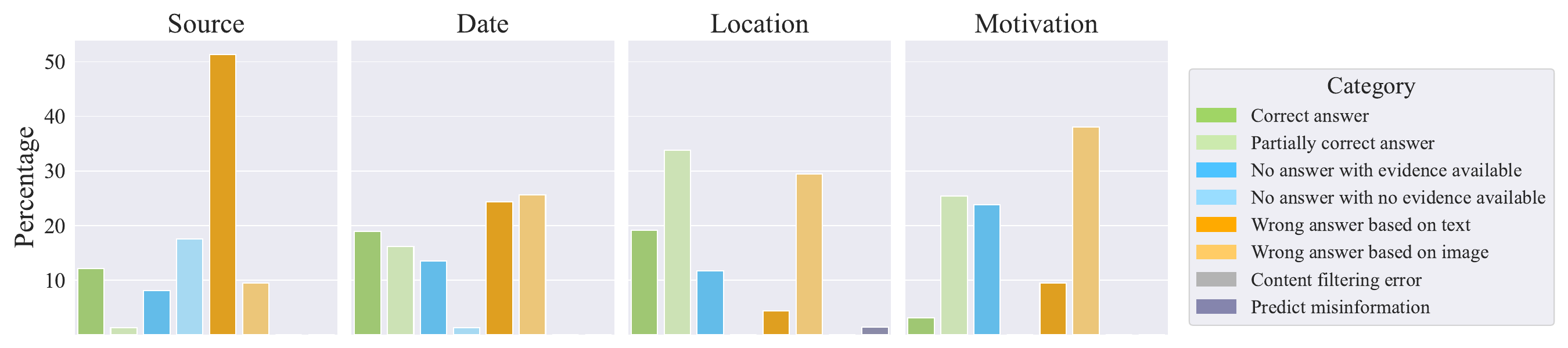}
    \caption{Qualitative analysis per pillar of multimodal zero-shot  Llava answers (\%).}
    \label{fig:qualitative_analysis_llava}
\end{figure*}

\begin{table*}
\centering
\resizebox{\textwidth}{!}{ %
\begin{tabular}{ccccccccccccccc}
\hline
\textbf{} &   \multicolumn{2}{c}{\textbf{Source}} & & \multicolumn{2}{c}{\textbf{Date}} & & \multicolumn{4}{c}{\textbf{Location}} &   & \multicolumn{3}{c}{\textbf{Motivation}} \\
\textbf{Area}  &  \textbf{RougeL} & \textbf{Meteor} & & \textbf{EM} & \textbf{$\Delta$} & & \textbf{RougeL} & \textbf{Meteor} &  \textbf{HL$\Delta$} &   \textbf{CO$\Delta$}  & & \textbf{RougeL} & \textbf{Meteor}  & \textbf{BertS} \\ 
 \hline
Africa & 3.8$_{\pm0.5}$  & 3.3$_{\pm0.4}$ & & 7.5$_{\pm0.0}$  & 23.3$_{\pm0.7}$ & & 28.9$_{\pm1.7}$ & 28.2$_{\pm1.4}$ &  32.4$_{\pm1.8}$ &  36.8$_{\pm2.0}$ & & 19.3$_{\pm0.3}$  & 15.0$_{\pm0.2}$  & 62.0$_{\pm0.1}$\\
Asia & 5.0$_{\pm1.3}$  & 4.4$_{\pm1.1}$  & & 8.0$_{\pm0.6}$  & 22.9$_{\pm1.0}$ & & 29.2$_{\pm0.4}$ & 27.7$_{\pm0.2}$ &  30.9$_{\pm0.3}$ &  36.5$_{\pm0.4}$ & & 19.1$_{\pm0.3}$ & 16.3$_{\pm1.3}$ & 64.2$_{\pm2.2}$\\
Europe  & 5.7$_{\pm1.4}$ & 5.1$_{\pm1.4}$ & & 7.2$_{\pm1.3}$ & 21.6$_{\pm2.2}$ & & 25.8$_{\pm1.5}$ & 24.9$_{\pm1.3}$ &  24.6$_{\pm2.4}$ &  30.9$_{\pm3.0}$ & & 20.6$_{\pm2.2}$  &  17.5$_{\pm2.2}$ & 68.1$_{\pm5.7}$ \\
North America & 6.1$_{\pm1.4}$ & 5.1$_{\pm1.2}$ & & 7.6$_{\pm1.3}$ & 21.5$_{\pm2.5}$ & & 32.1$_{\pm1.5}$ & 30.1$_{\pm2.0}$&  30.8$_{\pm2.5}$ & 32.7$_{\pm2.9}$ & & 21.3$_{\pm2.3}$  &  18.5$_{\pm2.6}$ & 69.4$_{\pm5.5}$\\
Oceania & 5.8$_{\pm1.4}$ & 4.7$_{\pm1.4}$ & & 6.1$_{\pm3.2}$ & 17.2$_{\pm8.9}$ & & 2.0$_{\pm0.0}$ & 5.5$_{\pm0.0}$&  18.6$_{\pm0.0}$ & 26.2$_{\pm0.0}$ & & 20.1$_{\pm3.3}$  &  17.2$_{\pm3.6}$ & 69.3$_{\pm5.5}$\\
 South America & 4.8$_{\pm2.5}$ & 3.9$_{\pm2.2}$ & & 5.1$_{\pm3.7}$ & 14.7$_{\pm9.9}$ & & 14.3$_{\pm0.0}$ & 14.5$_{\pm0.0}$&  24.8$_{\pm0.0}$ & 18.6$_{\pm0.0}$ & & 21.1$_{\pm3.7}$  &  16.7$_{\pm3.3}$ & 66.7$_{\pm7.7}$\\
\hline

\end{tabular}}

\caption{Multimodal zero-shot GPT4 results on the test set distributed by geographic area (\%). Reported results are averages over 3 iterations with standard deviations.}
\label{tab:results_geographic}
\end{table*}

We break down the qualitative analysis of zero-shot multimodal GPT4 and Llava per pillar in Figures \ref{fig:qualitative_analysis} and \ref{fig:qualitative_analysis_llava}, respectively. With GPT4, Location and Motivation follow similar distributions, containing most (partially) correct answers. They also concentrate all wrong answers based on the image content. Unlike Location, an answer is almost always provided for Motivation.
For Source, GPT4 tends to provide incorrect answers based on incorrect text evidence. Comparatively, it refrains more often from answering for Date, even when the image and the text evidence provide cues about the correct time period.\\
In the absence of relevant text evidence, GPT4 provides an answer based on the image content for Location more often than for Date. This is counter-intuitive, as one would expect the visual content to provide some cues for the Location of the image, of course, but also for its Date, for example, by recognizing celebrities or identifying the conflict or event depicted based on flags and clothes. This apparently weaker temporal knowledge of GPT4, compared to its geographical knowledge, deserves further investigation in future work.\\
Llava does not often abstain, resulting in a higher number of wrong answers based on text and image content, compared to GPT4. Many answers for Date, Location, and Motivation are partially correct, as they do not match the level of detail expected by the task. For example, only the year or country is provided, while the ground truth is more specific. This issue is partially corrected by adding demonstrations, as reflected in Table \ref{tab:results}.

\section{Results by geographic area}
\label{sec:geographic_results}

Table \ref{tab:results_geographic} decomposes the results obtained with multimodal zero-shot GPT4 by geographic area. We use the GeoNames hierarchy derived from the Location pillar to assign, when available, each image to its respective area. As shown in Figure \ref{fig:world-map}, Asia and Africa are well represented in 5Pils, while only a few instances have Location labels in South America and Oceania. The results in Table \ref{tab:results_geographic} reveal moderate differences between geographic areas in terms of average performance. One outlier is the Location prediction task for Oceania and South America, where the performances are lower than other areas by a large extent. For Date and Motivation prediction, the standard deviations are much higher in under-represented regions.

\end{document}